\documentclass[preprint,12pt]{elsarticle}

\usepackage[utf8]{inputenc} 
\usepackage{amsmath,amssymb}
\usepackage[most]{tcolorbox}
\usepackage{multicol}
\usepackage{graphicx}
\usepackage{float}
\usepackage{subcaption}
\usepackage{lscape}
\usepackage{multicol}
\usepackage[most]{tcolorbox} 
\usepackage[most]{tcolorbox} 
\usepackage{url}
\usepackage{hyperref}
\hypersetup{
  hidelinks
}

\journal{Energy}

\makeatletter
\def\@makefntext#1{%
  \noindent\makebox[0pt][r]{\@thefnmark.\ }#1}
\makeatother

\begin{document}

\begin{frontmatter}

\title{Capacity-constrained demand response in smart grids using deep reinforcement learning}

\author[inst1]{Shafagh Abband Pashaki}
\ead{spashaki@lincoln.ac.uk}

\author[inst1]{Sepehr Maleki}
\ead{smaleki@lincoln.ac.uk}

\author[inst1]{Amir Badiee}
\ead{abadiee@lincoln.ac.uk}

\affiliation[inst1]{
  organization={Lincoln AI Lab, University of Lincoln},
  city={Lincoln},
  country={United Kingdom}
}

\begin{abstract}
This paper presents a capacity-constrained incentive-based demand response approach for residential smart grids. It aims to maintain electricity grid capacity limits and prevent congestion by financially incentivising end users to reduce or shift their energy consumption. The proposed framework adopts a hierarchical architecture in which a service provider adjusts hourly incentive rates based on wholesale electricity prices and aggregated residential load. The financial interests of both the service provider and end users are explicitly considered. A deep reinforcement learning approach is employed to learn optimal real-time
incentive rates under explicit capacity constraints. Heterogeneous user preferences are modelled through appliance-level home energy management
systems and dissatisfaction costs. Using real-world residential electricity consumption and price data from three households, simulation results show that the proposed approach effectively reduces peak demand and smooths the aggregated load profile. This leads to an approximately $22.82\%$ reduction in the peak-to-average ratio compared to the no-demand-response case.

\end{abstract}


\end{frontmatter}

\section{Introduction}

\noindent Demand Response (DR) is a mechanism used by suppliers to manage real-time fluctuations in electricity networks. In such schemes, suppliers offer incentives to consumers to manage their power consumption during peak periods~\cite{ alfaverh2020demand}. This requires enabling consumers to lower their energy usage by reducing the load, shifting their consumption over time, or generating and storing energy at certain times~\cite{DOE2006_DR_benefits}. In modern smart grids, advanced Information and Communication Technology (ICT)
and Advanced Metering Infrastructure (AMI) enable bidirectional communication
between power suppliers and consumers. As a result, DR has
emerged as an important mechanism for improving the reliability and
sustainability of power systems~\cite{yu2016supply,shewale2020overview}.
 Despite its potential benefits, DR plays a limited role in most countries due to lack of trust, concerns about risk, and the complexity of participation~\cite{parrish2020systematic}. As a result, electricity supply and demand are balanced by ensuring that generation, reserves and network capacity are sufficient to meet demand~\cite{parrish2019demand}. A well-designed DR scheme can improve the balance between electricity supply and demand, enhance grid efficiency, and increase system flexibility, particularly in modern smart grids. Traditionally, the balance between supply and demand in the electricity network was managed by building enough power plants and infrastructure. In contrast, DR programmes help reduce supply costs by reducing the need for expensive new power plants or infrastructure that would only be used during peak load situations~\cite{oh2023multi}.

\medskip
\noindent DR methods are broadly classified into two main categories, namely price-based demand response (PBDR) and incentive-based demand response (IBDR)~\cite{shen2014role}. In PBDR, consumers adjust their electricity consumption in response to time-varying electricity prices~\cite{yan2018review}. IBDR programmes, on the other hand, provide fixed or time-varying incentives to customers to change or reduce their energy consumption during peak periods~\cite{ghasemkhani2019learning, van2023marl}. IBDR programmes are more dispatchable and less risky with respect to delivered flexibility compared to PBDR programmes, as power suppliers can more directly aim for and achieve a specific target level of load reduction via the incentive mechanism~\cite{eissa2018first}. Besides, it has been recognised that IBDR programmes are generally viewed as reward-based programmes in practice and exhibit more enduring and stable participant engagement, leading to more sustained benefits for consumers~\cite{li2016real}. Approximately 93\% of peak load reductions in the United States are achieved through IBDR programmes~\cite{asadinejad2017optimal,eslaminia2023enhancing}.

\medskip
\noindent DR programmes can be implemented in residential, commercial, and industrial establishments in a very similar manner~\cite{bakare2023comprehensive}. The residential sector, which accounts for more than 40\% of the total electricity demand, significantly impacts the electricity network and contributes to 30\% of greenhouse gas emissions~\cite{pallonetto2020assessment}. A DR scheme can flexibly manage residential loads under varying circumstances and can also incorporate strategies such as load profile management, strategy-driven conservation measures, and adjustments to consumer segments and market participation levels~\cite{panda2022residential}. Residential users participate and respond to DR programmes with varying levels of engagement across households. Financial and environmental benefits are commonly associated with residential participation in DR, with financial incentives often attracting greater attention. Nevertheless, concerns related to privacy, autonomy, and the perceived clarity of benefits may affect user trust and engagement~\cite{parrish2020systematic}. Transparent mechanisms, clear communication, and the involvement of trusted local actors have been identified as important elements for fostering participation in residential DR programmes. More broadly, residential DR is more likely to be adopted when the resulting economic savings outweigh perceived discomfort or loss of convenience, as many consumers exhibit limited willingness to compromise comfort for lower electricity costs~\cite{parrish2019demand}.

\medskip
\noindent
\textbf{Challenges.}
Despite the extensive development of DR programmes, designing effective real-time IBDR for residential smart grids remains challenging. In practice, residential DR must simultaneously (i) respect explicit grid capacity limits to prevent congestion, (ii) adapt to highly dynamic and uncertain electricity prices and consumption patterns, and (iii) account for heterogeneous end-user (EU) comfort preferences at the appliance level. Traditional approaches often rely on static or day-ahead formulations and simplified aggregate demand models, which limit their ability to respond to fast-changing operating conditions and to accurately represent household-level flexibility. As a result, many existing DR schemes struggle to deliver reliable peak reduction while maintaining sustained residential participation and acceptable EU comfort.

\medskip
\noindent\textbf{Existing solutions and limitations.}
Existing solutions to IBDR largely fall into two categories. On the one hand, model-based approaches such as mathematical optimisation, game-theoretic, and elasticity-based methods have been widely studied. While these methods provide analytical insight, they require strong modelling assumptions, accurate parameter estimation, and often lack scalability or real-time applicability in realistic residential settings. On the other hand, recent reinforcement learning (RL)-based approaches have demonstrated promising performance in learning adaptive incentive-setting or scheduling strategies from data. However, in many existing RL-based studies, EU response is modelled at an aggregated level, appliance-level flexibility is treated separately from incentive design, or grid capacity constraints are handled implicitly or under simplified assumptions. Consequently, there remains a gap between data-driven incentive learning and practical, capacity-aware residential DR with detailed device-level modelling.

\medskip
\noindent\textbf{Our contributions.}
To address these challenges, this paper proposes a capacity-constrained reinforcement learning-based demand response (CCRL-DR) framework for residential smart grids. The proposed approach explicitly integrates grid capacity limits into the RL state and reward formulation, enabling real-time, capacity-aware incentive design. A hierarchical architecture is considered, in which a service provider (SP)
learns optimal hourly incentive rates based on wholesale electricity prices and
aggregated residential demand. Heterogeneous EU preferences are
modelled through appliance-level home energy management systems (HEMSs)
and dissatisfaction costs. By combining deep RL with device-level response modelling and real-world load and price data, the proposed framework achieves effective peak shaving and load smoothing without inducing rebound peaks. Comparative simulations against an elasticity-based load reduction (EBLR) benchmark demonstrate that CCRL-DR delivers superior peak-to-average ratio (PAR) reduction while maintaining EU comfort and economic viability for both the SP and the EUs.

\section{Literature Review}
\noindent
Until now, a great deal of work has been devoted to the design of IBDR mechanisms for smart grids. These studies mainly rely on model-based formulations, including coupon/credit schemes, game-theoretic designs, and optimisation-based scheduling. The study in~\cite{fang2015coupon} presents a coupon-based DR
strategic bidding model for a load-serving entity under wind uncertainty. The
problem is formulated as a bi-level stochastic optimisation and reformulated
as a mixed-integer linear programming (MILP) problem to maximise the
load-serving entity’s profit. In a similar work, Li et al.~\cite{li2016dynamic} develop a dynamic coupon-based
IBDR programme for distributed energy systems with multiple load aggregators.
Coupons incentivise EUs to reduce peak-period demand and minimise operating
cost, while a fairness function ensures equitable incentive distribution
across aggregators. Erdinc et al.~\cite{erdinc2018novel} propose a credits-based incentive mechanism
for EUs to reduce critical load demand and maintain the balance between supply
and demand during peak periods. Ambient temperature uncertainty is
incorporated within a stochastic day-ahead framework. The studies~\cite{yu2017incentive,yu2018incentive} formulate grid operator-centric IBDR as a three-level hierarchy comprising a grid operator (GO), multiple SPs, and their EUs. The GO first announces an incentive to SPs, who then run EU-facing programmes and offer incentives to enrolled EUs to agree on specific demand-reduction amounts. In view of this hierarchical decision-making structure, a Stackelberg game is adopted to capture the interactions among these different participants. In~\cite{shahryari2018improved}, an improved IBDR is presented in which the elasticity is modelled as a function of price, time (hour), and EU type. The proposed IBDR can participate in both day-ahead and intra-day markets with time-varying incentive rates. Similarly, the work in~\cite{asadinejad2018evaluation} estimates incentive-based elasticity at the appliance level and shows that calculating elasticity at this level is more informative for operational decision-making than feeder-level value aggregated elasticity.

\medskip
\noindent
Beyond market design and incentive formulation, a number of studies have focused on incorporating uncertainty, risk management, and system-level operational constraints into IBDR programmes. In~\cite{ghazvini2015incentive}, a risk management scheme for the SP is presented
to address uncertainties in the day-ahead market and mitigate financial
losses. This is achieved by combining an IBDR programme for EUs with
distributed generation and energy storage systems. In a similar work,~\cite{ghazvini2015multi} proposes a multi-objective short-term IBDR scheduling model to manage market price risk and load uncertainty. In this framework, DR scheduling is performed simultaneously with the dispatch of generation and storage units while accounting for grid-imposed capacity obligations (e.g., capacity charges/penalties) and the resulting optimisation problem is solved using the non-dominated sorting genetic algorithm II (NSGA-II). In~\cite{vuelvas2018limiting}, a contract-based IBDR scheme between an
aggregator and a single EU is developed to induce individual rationality and
asymptotic incentive compatibility. Under this scheme, the EU self-reports
its baseline and reduction capacity. In more advanced classical studies, linear and nonlinear modelling for the IBDR programmes is implemented in four different power markets~\cite{rahmani2016modeling}. It revealed that the DR programmes can shift part of the electricity demand and save energy, while may lead to different demand pattern with different responsive load behavioural model. In~\cite{rahmani2016nonlinear}, a nonlinear IBDR is used in unit commitment problem by considering the nonlinear behaviours of residential EUs.

\medskip
\noindent However, most of the above-mentioned efforts on IBDR rely on traditional model-based methods, including stochastic programming, game-theoretic formulations, and MILP. These approaches typically require explicit system modelling and the identification of parameters that are dependent on EU characteristics, which can be difficult to obtain in practice. Moreover, model-based formulations often rely on simplifying assumptions that may limit their applicability in real-world residential settings. Although some studies incorporate uncertainty or capacity-related terms, these aspects are generally handled within offline or model-based frameworks. In addition, several existing works focus on a single EU or assume homogeneous users, thereby overlooking the impact of multiple EUs with heterogeneous characteristics. Furthermore, a large portion of the literature investigates DR in the context of day-ahead markets, while real-time operation is considered under more restrictive assumptions.

\medskip
\noindent
In recent years, RL has been increasingly adopted as an AI-based approach for sequential decision-making, in which an agent learns a policy through interaction with an environment to maximise cumulative reward~\cite{vazquez2019reinforcement}. In the context of smart energy management, RL-based methods have been applied to a variety of decision-making problems, including energy scheduling and incentive design~\cite{salazar2023reinforcement}. For example, in~\cite{costanzo2016experimental}, a model-assisted batch RL is
applied to building climate control under dynamic pricing. The sequential
decision-making problem is formulated as a Markov Decision Process (MDP), and a control policy is
learned via fitted Q-iteration. In addition, Mahapatra et al.~\cite{mahapatra2017energy} develop an advanced
neural fitted Q-learning approach to manage home appliances. The approach
aims to reduce peak-period power load, promote energy conservation, and
reduce the carbon footprint of residential dwellings. In one application~\cite{alfaverh2020demand}, a single-agent Q-learning-based
HEMS approach is proposed. Human preferences are considered by directly
integrating user feedback into the control logic using fuzzy reasoning to
shape the reward, enabling appliance scheduling through peak-to-off-peak
load shifting. In~\cite{kansal2024optimizing}, a Deep Reinforcement Learning (DRL) approach is proposed for home energy management, to optimise Photovoltaic (PV) self-consumption, focusing on domestic heating and hot-water systems. In a related study~\cite{ren2024data}, a data-driven DRL-based HEMS optimisation framework is developed to handle uncertain household parameters and high-dimensional, time-coupled decision-making. The methodology integrates a BiGRU-NN prediction model for forecasting PV output and electricity prices with a Soft Actor-Critic (SOC) agent to derive control policies and optimise scheduling decisions. In another approach to home energy management~\cite{xu2020multi}, this method successfully schedules diverse load types, including non-shiftable, power-shiftable, time-shiftable appliances, and EVs, ensuring a balance between operational efficiency and EU comfort. Similarly, the authors of~\cite{dusparic2015maximizing} present a decentralised multi-agent learning-based residential DR framework to better align household consumption with variable wind availability, thereby improving renewable integration in smart homes.

\medskip
\noindent
While the above studies mainly focus on household-level energy management, a separate line of research applies RL to IBDR, where decision-making is performed at the system or SP level. In one study, a single-agent RL framework is used to compute optimal incentive
rates for heterogeneous EUs, where the trade-off between EU incentive income
and dissatisfaction is jointly balanced with the SP’s objective~\cite{lu2019incentive}. Building on this formulation,~\cite{wen2020modified}
extends the same incentive-based RL framework by incorporating PV generation,
while~\cite{xu2020modified} further augments the model by including historical
incentive information in the decision-making process. In~\cite{van2023marl}, a decentralised Multi-Agent Reinforcement Learning (MARL) approach is applied to an IBDR
programme. A uniform incentive is provided to all residential EUs to reduce
energy consumption while maintaining grid capacity constraints and
preventing congestion. The incentive mechanism encourages heterogeneous EUs to participate while preserving their privacy. A novel disjunctively constrained knapsack problem optimization is incorporated to curtail or shift household appliance consumption based on the selected demand reduction. In~\cite{chung2020distributed}, the authors propose a model-free framework for
multi-household appliance scheduling formulated as a noncooperative
stochastic game. Distributed DRL is applied to search for the Nash
equilibrium consumption while preserving household privacy. In~\cite{zhang2019cooperative}, a cooperative multi-household MARL approach is used to schedule controllable appliances to minimize total energy costs while enforcing global physical constraints. Both approaches emphasize the ability to scale with the number of EUs and operate in real time.

\medskip
\noindent Overall, these studies demonstrate that RL can learn sequential decision policies through interaction with uncertain environments and balance cost objectives with EU comfort considerations. In the closest RL-based works on residential IBDR, consumer response is often
modelled at an aggregated EU level. Appliance-level modelling is typically
studied in HEMS settings without tight coupling to incentive-setting
mechanisms. In addition, although some RL formulations incorporate uncertainty or capacity-related terms, these aspects are typically handled implicitly or under simplified settings, particularly with respect to scalable and real-time operation under physical grid constraints.

\section{Demand response model}

\begin{figure}[!t]
    \centering
    \includegraphics[width=0.85\textwidth]{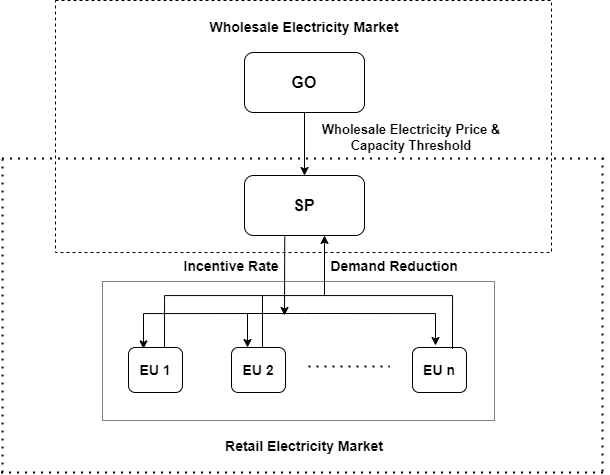}
    \caption{Hierarchical electricity market model.}
    \label{fig:Hierarchical electricity market model}
\end{figure}

\noindent In this section, we describe the system model and the interaction between the GO, SP, and residential EUs as shown in Fig.~\ref{fig:Hierarchical electricity market model}. The GO interacts with the wholesale market and announces both the hourly wholesale price and a capacity limit for the aggregated residential load. The SP purchases energy at the wholesale price and manages the demand of the participating EUs at the retail level. The SP communicates hourly financial incentive rates to the EUs, while the resulting demand adjustments are aggregated at the SP level. The aggregated load is constrained to remain below or close to the capacity limit, while EU discomfort is modelled through explicit dissatisfaction costs. 

\subsection{Service provider model}
\noindent The SP is the main aggregator and decision maker in the proposed IBDR
programme. For each hour $h$, the SP observes the aggregated demand
of the residential EUs and how close it is to the contractual capacity limit.
Taking into account the wholesale price announced by the GO, the SP then
sets hourly financial incentives to encourage EUs to reduce or shift part of
their flexible demand when the aggregated demand approaches this
limit. From the SP’s perspective, these demand reductions provide
demand-side flexibility by lowering the amount of energy that must be purchased from the
wholesale market; therefore, the SP’s performance is evaluated using the profit function in Eq.~\eqref{eq:SP_profit}.

\begin{equation}
\max \sum_{h=1}^H \sum_{n=1}^N 
\bigl( p_h \,\Delta E_{n,h} - \lambda_{n,h}\,\Delta E_{n,h} \bigr),
\label{eq:SP_profit}
\end{equation}

\begin{equation}
    \lambda_{\min} \le \lambda_{n,h} \le \lambda_{\max}, \qquad \forall n,h,
    \label{eq:SP_lam_bounds}
\end{equation}

\medskip
\noindent
where $n = 1,\dots,N$ denotes the index of EUs and $h = 1,\dots,H$ is the hourly time index. The term $p_h$ is the wholesale price at hour $h$, $\Delta E_{n,h}$ is the effective demand reduction of EU $n$ at hour $h$, and $\lambda_{n,h}$ is the incentive rate offered to that EU. The product $p_h \Delta E_{n,h}$ represents the saving in wholesale energy cost due to the reduction, whereas $\lambda_{n,h} \Delta E_{n,h}$ corresponds to the incentive payment~\cite{lu2019incentive,wen2020modified}. The bounds $\lambda_{\min}$ and $\lambda_{\max}$ specify an allowed range for the incentive rates, which can be interpreted as contractual or regulatory limits, and keep the offered incentives within a reasonable and practical level for both the SP and the EUs ~\cite{fang2015coupon,yu2017incentive}.

\subsection{End user model}

\noindent In the DR programme, each residential EU is assumed to be equipped with a HEMS, where applicable, that adjusts part
of its flexible energy demand in response to financial incentives proposed by
the SP. By participating in the programme, the EU can obtain a monetary reward
when reducing its demand, at the cost of modifying its preferred consumption
pattern. In each decision period, the HEMS determines the amount of flexible
demand to curtail or shift in time. Accordingly, the objective of each EU can be
written as

\begin{equation}
    \max \sum_{h=1}^{H}
    \Bigl[
        \rho\,\lambda_{n,h}\,\Delta E_{n,h}
        -
        (1-\rho)\,C^{\mathrm{dis}}_{n,h}
    \Bigr].
    \label{eq:EU_utility}
\end{equation}

\noindent The first term in Eq.\eqref{eq:EU_utility} represents the EU’s financial benefit from participating in the DR programme, while the second term corresponds to the dissatisfaction cost. Here, $\rho \in [0,1]$ is a weighting factor that reflects the trade-off between financial benefit and comfort. Larger values of $\rho$ correspond to EUs that are more responsive to financial incentives, whereas smaller values indicate EUs that place more emphasis on maintaining their comfort level. Similar to \cite{lu2019demand,van2023marl}, $C^{\mathrm{dis}}_{n,h}$ denotes the dissatisfaction cost of the $n$-th EU in hour $h$. 

\medskip
\noindent In each household $n$, the appliance set $\mathcal{D}_n$ is divided into three subsets,
time-shiftable appliances $\mathcal{D}_n^{\mathrm{TS}}$, power-controllable appliances $\mathcal{D}_n^{\mathrm{PC}}$ (e.g., air
conditioner), and non-shiftable appliances $\mathcal{D}_n^{\mathrm{NS}}$, which do not
participate in the DR programme and therefore remain unchanged, such that
$\mathcal{D}_n = \mathcal{D}_n^{\mathrm{TS}} \cup \mathcal{D}_n^{\mathrm{PC}} \cup
\mathcal{D}_n^{\mathrm{NS}}$. In this work, time-shiftable appliances are classified as either interruptible or non-interruptible. Interruptible time-shiftable appliances $\mathcal{D}_n^{\mathrm{TS,I}}$ (e.g., EV charging) can be paused and resumed in later hours, subject to meeting the daily energy requirement and the completion deadline.
 In contrast, non-interruptible time-shiftable appliances $\mathcal{D}_n^{\mathrm{TS,NI}}$ (e.g., washing machine, dishwasher, and dryer) must operate as a continuous block once started.
The total flexible energy reduction of EU $n$ in hour $h$ is then obtained by aggregating the contributions of all flexible appliances,
\begin{equation}
    \Delta E_{n,h}
    =
    \sum_{a \in \mathcal{D}_n^{\mathrm{PC}}}
        \Delta E^{\mathrm{PC}}_{n,a,h}
    +
    \sum_{a \in \mathcal{D}_n^{\mathrm{TS,I}}}
        \Delta E^{\mathrm{TS,I}}_{n,a,h}
    +
    \sum_{a \in \mathcal{D}_n^{\mathrm{TS,NI}}}
        \Delta E^{\mathrm{TS,NI}}_{n,a,h},
    \label{eq:EU_deltaE}
\end{equation}

\noindent where $\Delta E^{\mathrm{PC}}_{n,a,h}$ denotes the energy demand reduction
obtained by curtailing the consumption of power-controllable appliance $a$ at
hour $h$. The term $\Delta E^{\mathrm{TS,I}}_{n,a,h}$ represents the flexible
energy reduction resulting from temporarily reducing or postponing the
operation of interruptible time-shiftable appliance $a$ (e.g., shifting EV
charging to later hours). Finally, $\Delta E^{\mathrm{TS,NI}}_{n,a,h}$ captures
the energy demand reduction obtained by rescheduling the indivisible
operating block of non-interruptible time-shiftable appliance $a$ away from
high-load hours. Whenever part of a time-shiftable appliance’s energy demand is moved from a peak hour to lower-load hours (or hours below the capacity limit), the energy removed from that peak hour is counted in the corresponding $\Delta E^{\mathrm{TS,I}}_{n,a,h}$ or $\Delta E^{\mathrm{TS,NI}}_{n,a,h}$, while the same amount reappears in later hours as additional actual energy consumption. In this way, $\Delta E_{n,h}$ summarises both the curtailment of energy demand from PC appliances and the removal/relocation of TS loads from congested hours~\cite{lu2019demand,van2023marl}. Accordingly, non-shiftable appliances $\mathcal{D}_n^{\mathrm{NS}}$ do not contribute to $\Delta E_{n,h}$.

\medskip
\noindent To capture heterogeneous comfort preferences, we assign each controllable appliance
$a \in \mathcal{D}_n^{\mathrm{PC}} \cup \mathcal{D}_n^{\mathrm{TS}}$ of EU $n$
an appliance-specific dissatisfaction coefficient $\beta_{n,a} > 0$, which quantifies
the residents' discomfort due to power curtailment or time shifting. In practice,
$\beta_{n,a}$ is a user-defined parameter that residents can update in their HEMS
according to their preferences~\cite{van2023marl}. For PC appliances, we use a finite set of discrete curtailment levels~\cite{fahrioglu2002using}. The curtailment level of appliance $a$ in household $n$ and hour $h$ is represented by an integer $q_{n,a,h} \in \{0,1,\dots,m_a\}$, where $m_a$ is the number of available levels for that appliance. In practice, the actual energy reduction is implemented as a fraction of $E_{n,a,h}$, and the same fraction is used inside the dissatisfaction cost. The disutility from curtailing a PC appliance is then written as
\begin{equation}
    C^{\mathrm{PC}}_{n,a,h}
    =
    \beta_{n,a}
    \left(
        \frac{q_{n,a,h}}{m_a}\,
        E_{n,a,h}
    \right)^{2},
    \label{eq:PC_cost}
\end{equation}
\noindent and the corresponding energy reduction at level $q_{n,a,h}$ is
\begin{equation}
    \Delta E^{\mathrm{PC}}_{n,a,h}(q_{n,a,h})
    =
    \frac{q_{n,a,h}}{m_a}\,
    E_{n,a,h}.
    \label{eq:PC_deltaE}
\end{equation}
For a given incentive rate $\lambda_{n,h}$, the HEMS selects the curtailment level that maximises the net utility of appliance $a$ in hour $h$,
\begin{equation}
    q^{\star}_{n,a,h}
    \in
    \arg\max_{q \in \{0,1,\dots,m_a\}}
    \Bigl[
        \lambda_{n,h}\,
        \Delta E^{\mathrm{PC}}_{n,a,h}(q)
        -
        C^{\mathrm{PC}}_{n,a,h}(q)
    \Bigr],
    \label{eq:PC_best_response}
\end{equation}
and we write $\Delta E^{\mathrm{PC}}_{n,a,h}$ in Eq.\eqref{eq:EU_deltaE} as a shorthand for $\Delta E^{\mathrm{PC}}_{n,a,h}(q^{\star}_{n,a,h})$. This discrete best-response formulation is consistent with the search over a finite set of curtailment rates implemented in the numerical model.

\medskip
\noindent
For TS appliances (both interruptible and non-interruptible), the dissatisfaction is assumed to increase with the delay between the requested and the actual starting time of the appliance. Let $\Delta T_{n,a,h}$ denote the accumulated delay (in hours or time steps) when appliance $a$ of EU $n$ is actually operated in hour $h$. The associated dissatisfaction cost is modelled as
\begin{equation}
    C^{\mathrm{TS}}_{n,a,h}
    =
    \beta_{n,a}\,\Delta T_{n,a,h}^{2},
    \label{eq:TS_cost}
\end{equation}
so that longer waiting times result in higher discomfort.

\medskip
\noindent
In the implementation, each non-interruptible TS appliance is represented as a continuous  operating block of fixed length. 
When an incentive is offered and shifting is allowed, the HEMS searches over a finite set of feasible alternative starting times that respect the appliance-specific completion deadline, while also ensuring that the capacity limit is not violated. Among these candidates, the block is moved to an hour where the aggregated load is relatively low, so that the original operation during peak hours is replaced by operation in less congested periods. Accordingly, $\Delta E^{\mathrm{TS,NI}}_{n,a,h}$ in Eq.\eqref{eq:EU_deltaE} represents the energy removed from the original peak hours.

\medskip
\noindent
For interruptible TS appliances, the daily energy requirement is distributed over a given charging window. The HEMS adjusts the hourly charging profile in response to the incentive signal by favouring hours with lower load and sufficient remaining capacity, while ensuring that the total required energy is delivered before the end of the window. Similarly, $\Delta E^{\mathrm{TS,I}}_{n,a,h}$ in Eq.\eqref{eq:EU_deltaE} captures temporary reductions or postponements of appliance operation during high-load hours.

\medskip
\noindent
We denote the total dissatisfaction cost of EU $n$ in hour $h$ by $C^{\mathrm{dis}}_{n,h}$:
\begin{equation}
    C^{\mathrm{dis}}_{n,h}
    =
    \sum_{a \in \mathcal{D}_n^{\mathrm{PC}}}
        C^{\mathrm{PC}}_{n,a,h}
    +
    \sum_{a \in \mathcal{D}_n^{\mathrm{TS}}}
        C^{\mathrm{TS}}_{n,a,h},
    \label{eq:EU_Cdis}
\end{equation}
which aggregates the discomfort caused by both power curtailment and time shifting across all controllable appliances of EU $n$. Substituting Eqs.\eqref{eq:EU_deltaE} and \eqref{eq:EU_Cdis} into
\eqref{eq:EU_utility} expresses the EU objective directly in terms of
appliance-level demand reductions and their dissatisfaction costs.

\subsection{Objective function}

\noindent
Considering the profits of both the SP and the EUs, the final objective function is presented as follows:
\begin{equation}
    \max
    \sum_{h=1}^{H}
    \Bigg\{
    \sum_{n=1}^{N}
    \Bigg[
        \underbrace{
            \big(p_h - \lambda_{n,h}\big)\,\Delta E_{n,h}
        }_{\text{SP's profit}}
        +
        \underbrace{
            \rho\,\lambda_{n,h}\,\Delta E_{n,h}
            - (1-\rho)\,C^{\mathrm{dis}}_{n,h}
        }_{\text{EU's profit}}
    \Bigg]
    + \Phi_h
    \Bigg\}.
    \label{eq:global_objective}
\end{equation}

\noindent In the numerical implementation, we augment the hourly reward $r_h$ with a shaping term $\Phi_h$ to encourage practically desirable behaviours while keeping the economic structure of Eq.~\eqref{eq:global_objective} unchanged. Specifically, $\Phi_h$ comprises four components; (i) a constant bonus of $+5$ when no reduction is required and no incentives are issued, (ii) a penalty of $-5\sum_{n=1}^{N}\lambda_{n,h}$ when no reduction is required but incentives are still sent, (iii) a penalty proportional to the unmet reduction, $-15\,R_h^{\mathrm{miss}}$, which is doubled if $\sum_{n=1}^{N}\lambda_{n,h}=0$ despite a positive required reduction, and (iv) an over-reduction penalty of $-0.5\,R_h^{\mathrm{over}}$ to discourage unnecessary curtailment beyond the required level. Here, $R_h^{\mathrm{miss}}$ and $R_h^{\mathrm{over}}$ denote the missing and excess reductions at hour $h$, respectively.

\section{Incentive optimisation}
\noindent In the
proposed IBDR scheme, the SP is modelled as
the RL agent. In this section, we model the decision-making problem as a finite-horizon MDP, define the state, action, and reward, and employ a Double Deep Q-Network (DDQN) algorithm to learn an incentive rate policy.

\subsection{Markov decision process for incentive-based demand response}
\label{sec:mdp}
\noindent
MDP has been widely adopted to study RL-based decision-making
problems. A MDP provides a
classical framework for sequential decision making, where the actions
taken by the agent influence not only the immediate reward, but also the
subsequent states and hence future rewards. It is characterised by the
Markov property, namely that the state transitions depend only on the
current state and current action, and are independent of all previous
states and actions. In the MDP model considered here, the key components
are a discrete hour index $h \in \{1,\dots,H\}$, a state
$s_h \in \mathcal{S}$, an action $a_h \in \mathcal{A}$ and a reward
$r_h = R(s_h,a_h)$. One episode of this MDP thus constitutes the finite
sequence $(s_1,a_1,r_1), (s_2,a_2,r_2), \dots, (s_H,a_H,r_H)$, where $H$
denotes the final hour in the decision horizon.

\begin{figure}[!t]
    \centering
    \includegraphics[width=0.85\textwidth]{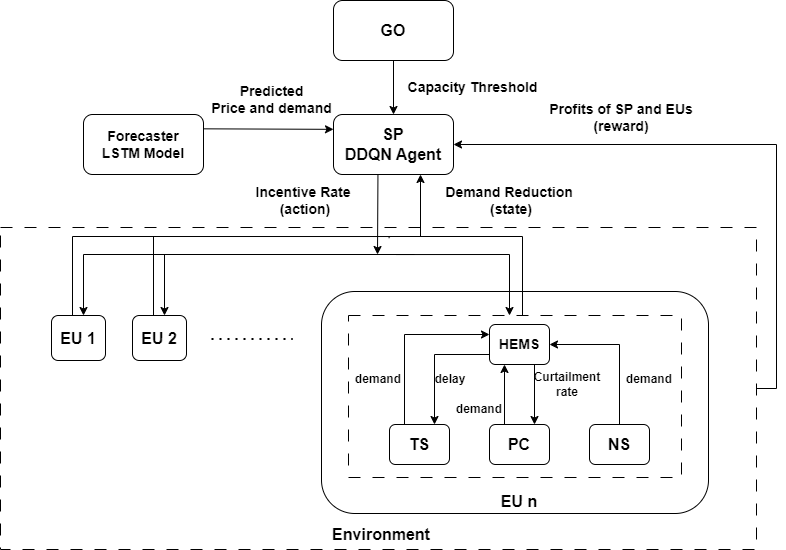}
    \caption{Schematic of RL to find the optimal incentive rates.}
    \label{fig:rl-incentive-arch}
\end{figure}

\medskip
\noindent
In line with the above MDP formulation, Fig.~\ref{fig:rl-incentive-arch} illustrates the RL-based decision-making structure adopted in
this work, where the SP serves as the agent and the participating
residential EUs are considered the environment. At each hour, the SP announces a vector of incentive rates as the action. In response, the residential EUs adjust their electricity consumption. The resulting aggregated load, electricity price, and required demand reduction with respect to the capacity limit define the system state and the per-hour reward $r_h$, as given in Eq.~\eqref{eq:reward_hour}. The reward $r_h$ corresponds to the hourly contribution of the global objective defined in Eq.~\eqref{eq:global_objective}.

\begin{equation}
r_h =
\sum_{n=1}^{N}
\Big[
    (p_h - \lambda_{n,h}) \Delta E_{n,h}
    + \rho \lambda_{n,h} \Delta E_{n,h}
    - (1-\rho) C^{\mathrm{dis}}_{n,h}
\Big]
+ \Phi_h.
\label{eq:reward_hour}
\end{equation}

\medskip
\noindent
Taking the long-term returns into account, the SP must consider not only
the current reward but also future rewards. The cumulative discounted
reward from hour $h$ is defined as
\begin{equation}
G_h
=
\sum_{k = h}^{H}
\gamma^{\,k-h} \, r_k,
\label{eq:discounted_return}
\end{equation}
where $\gamma \in [0,1]$ is the discount factor that controls the
relative importance of future versus present rewards.
Under a given policy $\pi$, the action--value function of taking action $a_h$
in state $s_h$ is defined as
\begin{equation}
Q^{\pi}(s_h, a_h)
=
\mathbb{E}_{\pi}\!\left[\, G_h \mid s_h, a_h \,\right].
\label{eq:q_policy}
\end{equation}

\noindent
The optimal value is then $Q^{*}(s_h, a_h) = \max\limits_{\pi} Q^{\pi}(s_h, a_h)$.
An optimal policy can then be derived by selecting, in each state, the action with the highest action–value. In
the following subsection, we describe how $Q^{\ast}(s_h,a_h)$ is
approximated using a DDQN algorithm.

\subsection{Double Deep Q-Network}
\noindent
In problems with a relatively high-dimensional and continuous state
space, a tabular representation of the action--value function $Q(s,a)$
is not practical. Instead, we approximate $Q(s,a)$ by a deep neural
network $Q(s,a;\theta)$ with parameters $\theta$, which takes the current
state $s_h$ as input and outputs one Q-value per discrete incentive
action \cite{mnih2015human}. Each discrete action corresponds to a
pre-defined vector of incentive rates. To improve learning stability and mitigate the overestimation bias of standard deep Q-networks (DQN), we adopt a DDQN architecture. The DDQN algorithm employs two neural networks, denoted by $Q$ and $Q^{-}$, which share the same network architecture but have different parameter sets. The parameters $\theta$ of the policy network $Q$ are used to select the action corresponding to the maximum Q-value, while the parameters $\theta^{-}$ of the target network $Q^{-}$ are used to evaluate the Q-value of the selected action \cite{xi2019novel,van2016deep}.

\medskip
\noindent
During training, transitions of the form
$(s_h,a_h,r_h,s_{h+1},\text{done}_h)$, where $\text{done}_h \in \{0,1\}$
indicates whether the episode has terminated at hour $h$, are stored in
a replay buffer $\mathcal{B}$.
Mini-batches are sampled uniformly from $\mathcal{B}$ to update the
policy network \cite{barto2021reinforcement}. For a sampled transition indexed by $j$, the DDQN
target is computed as
\begin{equation}
y_j
=
r_j
+
\gamma \bigl(1 - \text{done}_j\bigr)\,
Q\Bigl(
s'_j,\;
\arg\max_{a'} Q(s'_j,a';\theta);\;
\theta^-
\Bigr),
\label{eq:ddqn_target}
\end{equation}
where $s'_j$ denotes the next state and $\gamma$ is the discount factor
introduced in Eq.\eqref{eq:discounted_return}. The parameters $\theta$ of the
policy network are then updated by minimising a squared-error (Huber)
loss between $y_j$ and the current estimate $Q(s_j,a_j;\theta)$ over the
mini-batch. The target-network parameters $\theta^-$
are updated more slowly using a soft update rule,
\begin{equation}
\theta^-
\leftarrow
\tau\,\theta
+
(1-\tau)\,\theta^-,
\label{eq:soft_update}
\end{equation}
with a small $\tau \in (0,1)$ specified in the configuration.

\subsection{Real-time algorithm with LSTM forecasts and DDQN}

\noindent
To implement the CCRL-DR framework in practice, we combine the LSTM-based forecasting models with the DDQN scheme described in the previous subsections. Algorithm~\ref{tab:algorithm_ccrl_dr} summarises the resulting real-time IBDR procedure. Each episode corresponds to a single day and consists of $H$ hourly decision steps. At the beginning of each episode, time-dependent input variables such as the day of the week, hour of the day, and seasonal indicators are constructed for all hours $h = 1,\dots,H$ and fed into the pre-trained LSTM models. These models then generate predictions of the wholesale electricity price $F_{\mathrm{uPrice}}$ and the aggregated residential load $F_{\mathrm{uLoad}}$ by rolling their one-step-ahead forecasts over the whole day. The forecasted values for each hour are later incorporated into the RL state, providing the agent with a forward-looking view of the expected operating conditions.

\medskip
\noindent
In the DDQN control loop, the environment is initialised with the forecasted
price and load trajectories. At each hour $h$, the SP agent observes the
current state $s_h$ (including forecasts, current load, and capacity
information) and selects an incentive rate vector $a_h = \{\lambda_{n,h}\}$
using an $\varepsilon$-greedy policy derived from the policy network. The environment applies these incentive rates, and the EUs adjust their
appliance-level consumption according to the HEMS models. The resulting
demand reductions and updated load profile yield the next state $s_{h+1}$
and the immediate reward $r_h$, based on the per-hour contribution of the
global objective in Eq.~\eqref{eq:global_objective}. The transition $(s_h,a_h,r_h,s_{h+1},\mathrm{done})$ is stored in the replay buffer, and when the buffer has enough samples, mini-batches are sampled to update the policy network parameters by minimising the DDQN loss with target values defined in Eq.~\eqref{eq:ddqn_target}. The target network parameters are softly updated towards the policy parameters using the rule in Eq.~\eqref{eq:soft_update}, and the exploration rate $\varepsilon$ is gradually decayed according to a predefined schedule. 

\begin{table}[H]
\centering
\caption{Algorithm for real-time IBDR with LSTM and DDQN}

\begin{tabular}{p{0.03\textwidth}p{0.9\textwidth}}
\hline
\multicolumn{2}{l}{\parbox{\linewidth}{\textbf{Inputs:} policy network $Q(s,a;\theta)$, target network $Q(s,a;\theta^{-})$, replay buffer  $\mathcal{B}$, weighting factor $\rho$, DDQN hyperparameters $(\gamma,\varepsilon,\tau,\alpha)$, incentive rate bounds $\lambda_{\min},\lambda_{\max}$.}}\\
\hline

1:  & \textbf{for} episode $e = 1,\dots,N_{\mathrm{ep}}$ \textbf{do} \\
2:  & \quad \textbf{\% LSTM-based price forecasting} \\
3:  & \quad \textbf{for} hour $h = 1,\dots,H$ \textbf{do} \\
4:  & \quad \quad DofWeek[$h$] $\leftarrow$ updateDayOfWeek($h$); HofDay[$h$] $\leftarrow$ updateHourOfDay($h$); MofYear[$h$] $\leftarrow$ updateMonthOfYear($h$); IsWeekend[$h$] $\leftarrow$ updateWeekendFlag($h$); IsHoliday[$h$] $\leftarrow$ updateHolidayFlag($h$) \\
5:  & \quad \quad PrSeq[$h$] $\leftarrow$ updateHistoricalPriceSequence() \\
6:  & \quad \quad FuPrice[$h$] $\leftarrow$ PriceLSTM(DofWeek[$h$], HofDay[$h$], MofYear[$h$], IsWeekend[$h$], IsHoliday[$h$], PrSeq[$h$]) \\
7:  & \quad \textbf{end for} \\
8:  & \quad \textbf{\% LSTM-based load forecasting} \\
9:  & \quad \textbf{for} hour $h = 1,\dots,H$ \textbf{do} \\
10: & \quad \quad LoSeq[$h$] $\leftarrow$ updateHistoricalLoadSequence() \\
11: & \quad \quad FuLoad[$h$] $\leftarrow$ LoadLSTM(DofWeek[$h$], HofDay[$h$], MofYear[$h$], IsWeekend[$h$], IsHoliday[$h$], LoSeq[$h$]) \\
12: & \quad \textbf{end for} \\
13: & \quad \textbf{\% DDQN-based incentive learning} \\
14: & \quad $s_1 \leftarrow \text{Environment.reset}(\text{FuPrice},\text{FuLoad})$ and obtain initial state $s_1$. \\
15: & \quad \textbf{for} decision step (hour) $h = 1,\dots,H$ \textbf{do} \\
16: & \quad \quad Using an $\varepsilon$-greedy policy derived from the policy network, select
$a_h=\{\lambda_{n,h}\}_{n=1}^{N}$, where $\lambda_{n,h}\in[\lambda_{\min},\lambda_{\max}]$.
 \\
17: & \quad \quad Environment applies $\lambda_{n,h}$, computes demand reductions and the total reward $r_h$, and returns next state $s_{h+1}$ and termination flag $\text{done} \in \{0,1\}$. \\
18: & \quad \quad Store $(s_h,a_h,r_h,s_{h+1},\text{done})$ in replay buffer $\mathcal{B}$. \\
19: & \quad \quad \textbf{if} $B$ has enough samples and an update step is due \textbf{then} \\
20: & \quad \quad \quad Sample a mini-batch $(s_j,a_j,r_j,s'_j,\text{done}_j)$ from $\mathcal{B}$. \\
21: & \quad \quad \quad Select the next greedy action using the policy network: \\
    & \quad \quad \quad $a^\star \leftarrow \arg\max_{a'} Q(s'_j,a';\theta)$. \\
22: & \quad \quad \quad Compute the DDQN target: $y_j = r_j + \gamma (1-done_j)\, Q(s'_j, a^\star; \theta^-)$.
 \\
23: & \quad \quad \quad Update $\theta$ to minimise the loss: $L = \mathbb{E}\big[(y_j - Q(s_j,a_j;\theta))^2\big]$. \\
24: & \quad \quad \quad Soft-update the target network: $\theta^{-} \leftarrow \tau \theta + (1-\tau)\theta^{-}$. \\
25: & \quad \quad \textbf{end if} \\
26: & \quad \quad Update $\varepsilon$ according to the exploration decay. \\
27: & \quad \quad \textbf{if} done \textbf{then} break. \\
28: & \quad \textbf{end for} \\
29: & \textbf{end for} \\
\hline
\end{tabular}
\label{tab:algorithm_ccrl_dr}
\end{table}

\section{Experimental analysis}

\noindent In this section, we present the results obtained by applying our proposed approach using real-world data. We evaluated the precision of our forecasting models and the effectiveness of our CCRL-DR scheme.

\subsection{Price and load forecasting}

\noindent We adopt LSTM-based DNN models for both price and load forecasting and use their one-step-ahead predictions as external inputs to the proposed IBDR scheme. Following the preprocessing methodology in~\cite{van2023marl}, we utilized publicly available residential load datasets from Dataport~\cite{dataport} and extracted hourly load data for three randomly selected residential EUs located in Austin, Texas, USA, over the period from 1 April 2018 to 30 September 2018. This interval covers the months with the highest residential electricity consumption and peak load patterns, providing a suitable context for evaluating DR strategies. The testing period for the DR experiments was set to 1–31 July 2018, and is kept distinct from the training set. In parallel, wholesale price data for the same period were collected from the ERCOT market via EnergyOnline~\cite{energyonline}, with training and testing splits aligned with those of the load data.

\medskip
\noindent
To generate the required forecasts, we developed two separate LSTM-based DNN
models. One model predicts the hourly wholesale electricity price, while
three identical models predict the load of each individual EU, and the
aggregated residential load is obtained by summing these forecasts. The two models share the same overall architecture but are trained independently on their respective historical time series. Adequate selection of input variables is critical for reliable performance~\cite{lu2019incentive,wen2020modified}. For both targets, the input feature vector at each time step includes time-related variables (month of year, day of week, hour of day, and binary indicators for holidays and weekends) together with recent historical values of the quantity to be forecast. Short-term dynamics are represented by several lagged observations from the previous hours, whereas longer-term patterns are captured by including values from the same hour on previous days. The full set of input features for the price and load forecasting models is summarised in Table~\ref{tab:rnn-gru-inputs}.

\begin{table}[!htbp]
\centering
\caption{Inputs to the LSTM-based forecasting models}
\label{tab:rnn-gru-inputs}
\begin{tabular}{ll|ll}
\hline
\multicolumn{2}{c|}{Inputs for price forecasting} & \multicolumn{2}{c}{Inputs for load forecasting} \\
Index & Description & Index & Description \\
\hline
1  & Month of year (1--12)        & 1  & Month of year (1--12)        \\
2  & Day of week (1--7)           & 2  & Day of week (1--7)           \\
3  & Hour of day (1--24)          & 3  & Hour of day (1--24)          \\
4  & Is holiday (0 or 1)          & 4  & Is holiday (0 or 1)          \\
5  & Is weekend (0 or 1)          & 5  & Is weekend (0 or 1)          \\
6  & Price of hour $h - 1$        & 6  & Load of hour $h - 1$         \\
7  & Price of hour $h - 2$        & 7  & Load of hour $h - 2$         \\
8  & Price of hour $h - 3$        & 8  & Load of hour $h - 3$         \\
9  & Price of hour $h - 24$       & 9  & Load of hour $h - 24$        \\
10 & Price of hour $h - 25$       & 10 & Load of hour $h - 25$        \\
11 & Price of hour $h - 26$       & 11 & Load of hour $h - 26$        \\
12 & Price of hour $h - 48$       & 12 & Load of hour $h - 48$        \\
13 & Price of hour $h - 49$       & 13 & Load of hour $h - 49$        \\
14 & Price of hour $h - 50$       & 14 & Load of hour $h - 50$        \\
\hline
\end{tabular}
\end{table}

\medskip
\noindent
Forecasting with the proposed LSTM-based models follows a standard supervised learning pipeline~\cite{choi2021deep}. Each training sample consists of an input sequence constructed from the features in Table~\ref{tab:rnn-gru-inputs} and the corresponding target value (future price or load at hour $h$). The network parameters are updated by backpropagation through time, using the mean squared error (MSE) between predicted and true values as the loss function and the Adam optimiser for gradient-based updates~\cite{abumohsen2023electrical}. The main hyper-parameters, including the number of LSTM layers, hidden units, dropout rate, window size and forecast horizon, are reported in Table~\ref{tab:hyperparams}. To generate the required forecasts, we developed two separate LSTM-based DNN
models. One model predicts the hourly wholesale electricity price, while
three identical models predict the load of each individual EU, and the
aggregated residential load is obtained by summing these forecasts. After training, the LSTM-based DNNs are kept fixed and, at each hour, map the most recent input sequence to one-step-ahead forecasts of the wholesale price and aggregated residential load~\cite{memarzadeh2021short}.

\begin{table}[t!]
\centering
\caption{Hyper-parameters of the LSTM-based forecasting models}
\label{tab:hyperparams}
\begin{tabular}{ll}
\hline
Hyper-parameter        & Value \\
\hline
Number of LSTM layers   & 2     \\
Hidden units            & 64    \\
Dropout                 & 0.2   \\
Window size             & 24    \\
Forecast horizon        & 1     \\
Optimiser               & Adam  \\
Loss function           & MSE   \\
\hline
\end{tabular}
\end{table}

\begin{figure}[!t]
    \centering
    \includegraphics[width=0.85\textwidth]{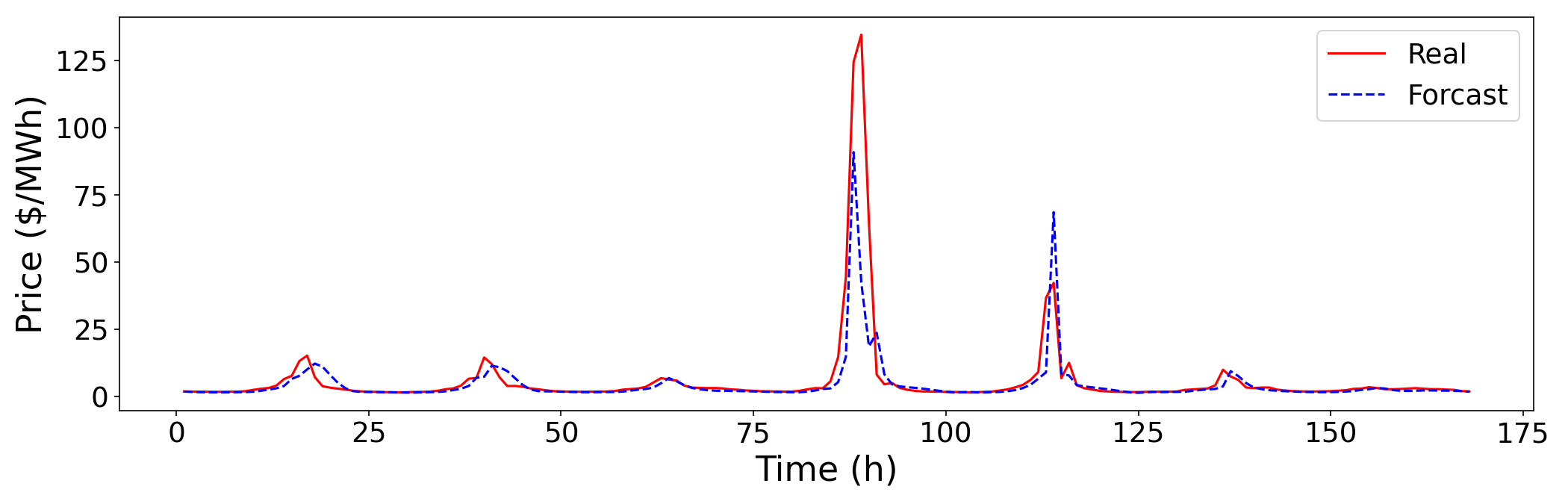}
    \caption{Wholesale price forecasting for July 23--29, 2018.}
    \label{fig:price-week4}
\end{figure}

\begin{figure}[H]
    \centering

    \begin{subfigure}[b]{0.85\textwidth}
        \centering
        \includegraphics[width=\textwidth]{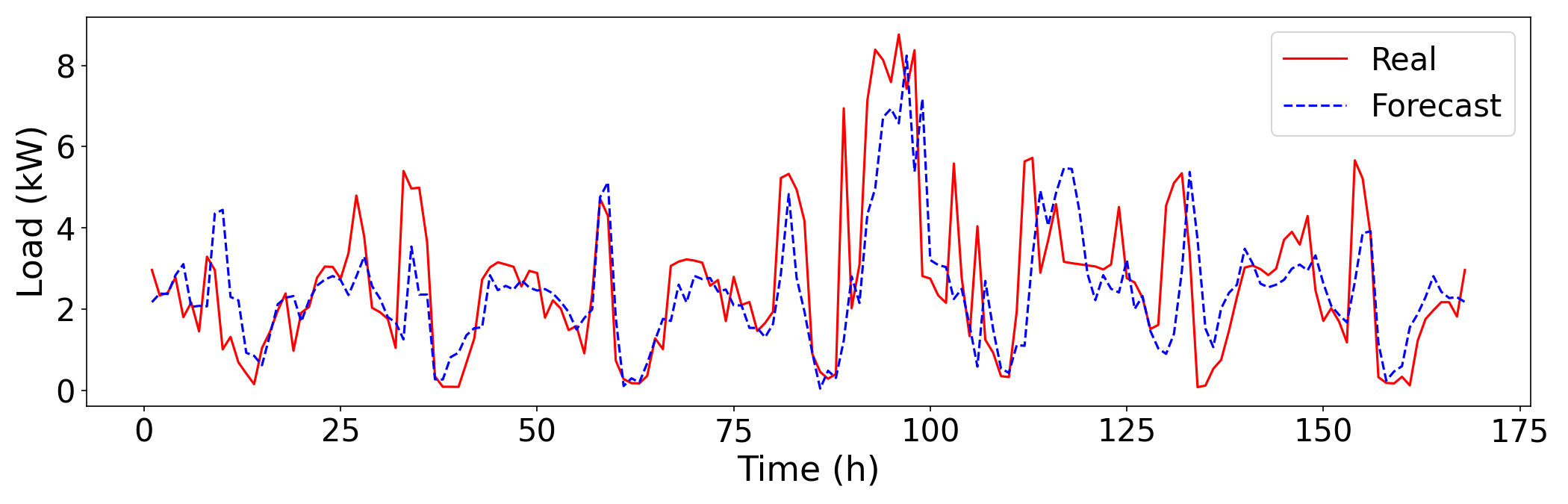}
        \caption{End User 1}
        \label{fig:load-house1}
    \end{subfigure}

    \vspace{0.35cm}

    \begin{subfigure}[b]{0.85\textwidth}
        \centering
        \includegraphics[width=\textwidth]{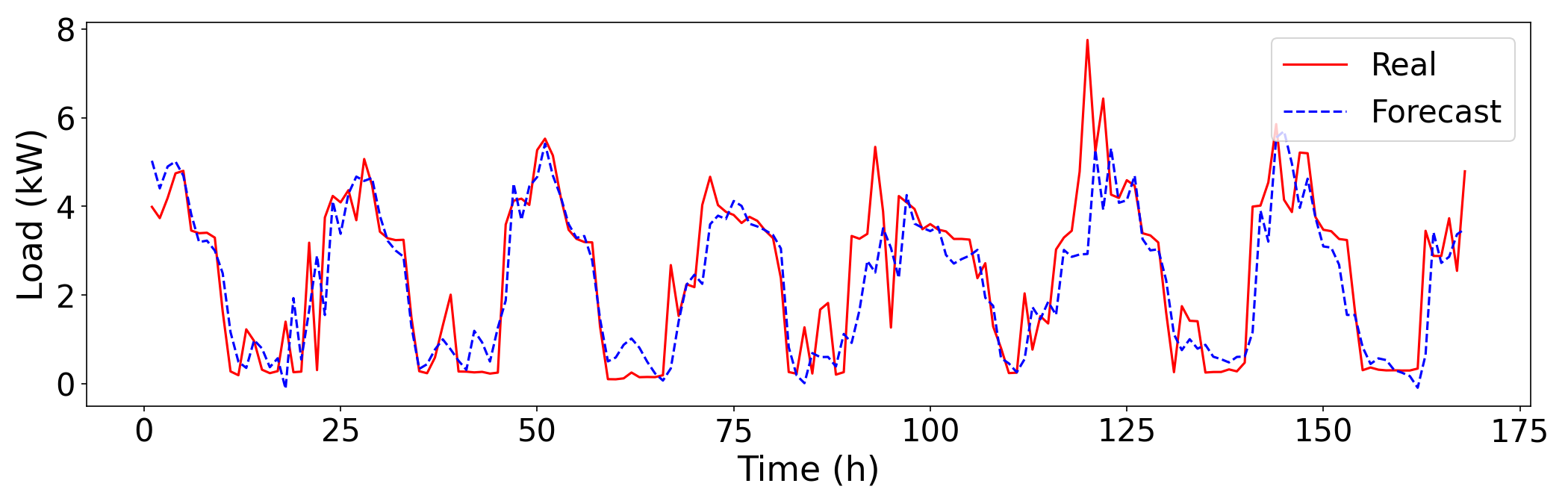}
        \caption{End User 2}
        \label{fig:load-house2}
    \end{subfigure}

    \vspace{0.35cm}

    \begin{subfigure}[b]{0.85\textwidth}
        \centering
        \includegraphics[width=\textwidth]{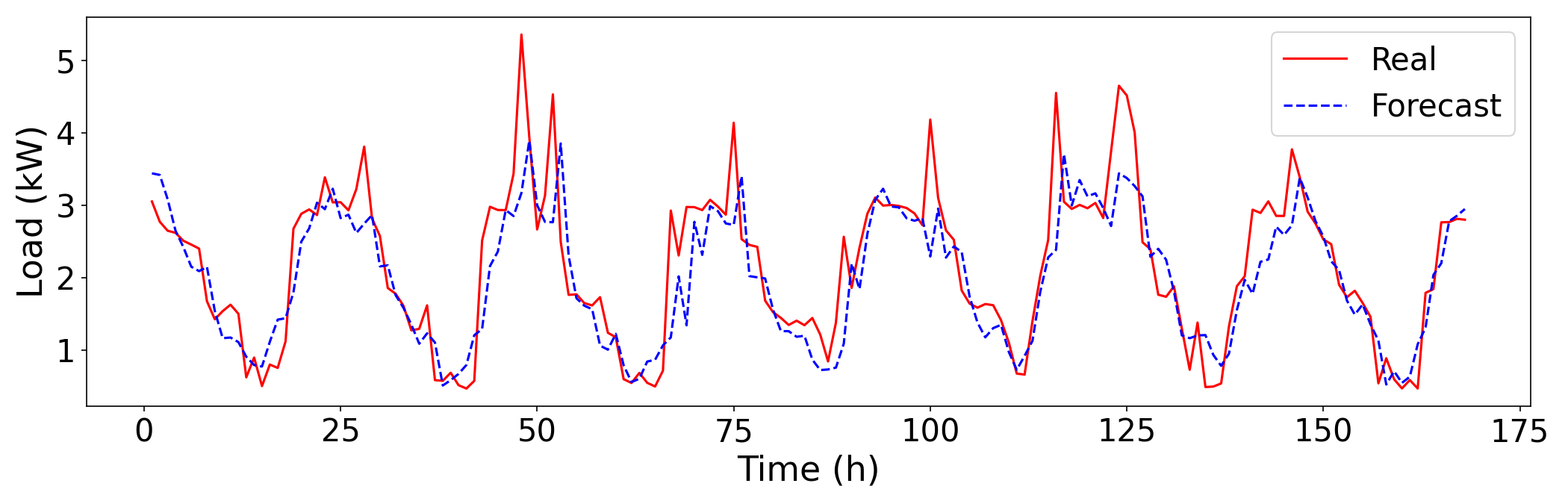}
        \caption{End User 3}
        \label{fig:load-house3}
    \end{subfigure}

    \caption{Load forecasting of the three end users for July 23--29, 2018.}
    \label{fig:load-forecast}
\end{figure}

\medskip
\noindent
Figs.~\ref{fig:price-week4} and~\ref{fig:load-forecast} compare the predicted and actual electricity prices and load demands during the final week of the testing period (23--29 July 2018). In these plots, the actual values are depicted by solid red lines, while the forecasted values are shown by blue dashed lines. The corresponding forecast errors, quantified in terms of Mean Absolute Error (MAE) and Mean Absolute Percentage Error (MAPE), are summarised in Table~\ref{tab:forecast-accuracy}. The relatively low MAE and MAPE values indicate good forecasting accuracy.
In addition, the adopted LSTM architecture closely tracks the main daily and
weekly patterns in both prices and loads, making it suitable for use within
the proposed CCRL-DR framework.

\begin{table}[H]
\centering
\caption{Accuracy results of the forecasting model}
\label{tab:forecast-accuracy}
\begin{tabular}{lcc}
\hline
 & MAE & MAPE \\
\hline
Wholesale price forecasting model & 1.87 & 0.53 \\
Load forecasting model of EU1     & 0.90 & 0.93 \\
Load forecasting model of EU2     & 0.72 & 0.78 \\
Load forecasting model of EU3     & 0.41 & 0.24 \\
\hline
\end{tabular}
\end{table}

\subsection{Evaluation}
\noindent
In this part, we describe how we applied our RL-based algorithm to manage electricity consumption using the forecasts generated by our price and load models. Specifically, we utilized the forecasted data for July 27, 2018, to assess the algorithm's effectiveness. The objective was to keep total electricity consumption within a predetermined capacity threshold while considering financial benefits for both the SP and EUs.\medskip

\noindent 

\subsubsection{Simulation Setup}

\noindent
For evaluating the performance of the proposed RL-based DR algorithm, we conduct simulations using a simplified scenario consisting of a single SP and three residential EUs. Each EU corresponds to a real household in the 2018 data set (house IDs 661, 3039, and 8565). In all simulations, the hourly load and price trajectories fed to the RL environment are the forecasted values obtained from our forecasting models. The aggregate capacity threshold for the three EUs is set to $75\%$ (approximately $7~\mathrm{kW}$) of the average daily peak of their total demand and is kept fixed throughout training and testing.

\medskip

\noindent
In the simulations, we use the DDQN-based framework to determine hourly incentive rates for each EU. The incentive bounds are set to $\lambda_{\min}=0$ and $\lambda_{\max}=0.95\,p_h$, such that the incentives are directly tied to the current market price and never exceed $95\%$ of $p_h$. To model the trade-off between incentive income and dissatisfaction cost on the EU side, we use a weighting factor $\rho$ in the EU-level reward; in all simulations we set $\rho = 0.9$ and assume it is identical for all EUs. At the appliance level, the dissatisfaction coefficients used in the EU reward model are specified per device type. We associate each appliance with a normal distribution and sample heterogeneous dissatisfaction coefficients across households during training \cite{van2023marl}. The corresponding mean and standard deviation values for each appliance type are reported in Table~\ref{tab:appliance_params} and are kept fixed across all experiments.

\begin{table}[H]
\centering
\caption{Household appliances and their dissatisfaction parameters (TS: time-shiftable, PC: power-controllable, NI: non-interruptible, I: interruptible).}
\label{tab:appliance_params}
\begin{tabular}{llll}
\hline
Appliance            & Type   & \multicolumn{2}{c}{Dissatisfaction coeff.} \\
                     &        & mean               & std                  \\
\hline
Dryer (DR)           & TS, NI & 0.10                        & 0.10 \\
Washing Machine (WM) & TS, NI & 0.40                        & 0.10 \\
Dish Washer (DW)      & TS, NI & 0.20                        & 0.10 \\
Electric Vehicle (EV)& TS, I  & 0.05                        & 0.10 \\
Air Conditioner (AC) & PC     & 3.50                        & 2.00 \\
\hline
\end{tabular}
\end{table}

\medskip

\noindent
The DDQN agent uses fully connected policy and target networks with two hidden layers of 128 and 64 ReLU units. Each training episode corresponds to one full day with $H = 24$ hourly time steps, and the agent is trained for $N_{\text{ep}} = 2500$ episodes. At the beginning of each episode, one training day from 2018 is sampled uniformly at random and the corresponding forecasted load and price trajectories are used as the environment input. The policy network is optimised with the Adam algorithm with learning rate $\alpha = 10^{-4}$ and discount factor $\gamma = 0.99$, using a replay buffer of size 50{,}000 and mini-batches of 256 transitions. We adopt an $\varepsilon$-greedy exploration strategy with $\varepsilon_{\text{start}} = 1.0$, $\varepsilon_{\min} = 0.01$, and exponential decay rate $0.998$. The target network parameters are updated using the soft-update rule in Eq.\eqref{eq:soft_update} with $\tau = 0.003$.

\begin{figure}[H]
    \centering
    \includegraphics[width=0.8\textwidth]{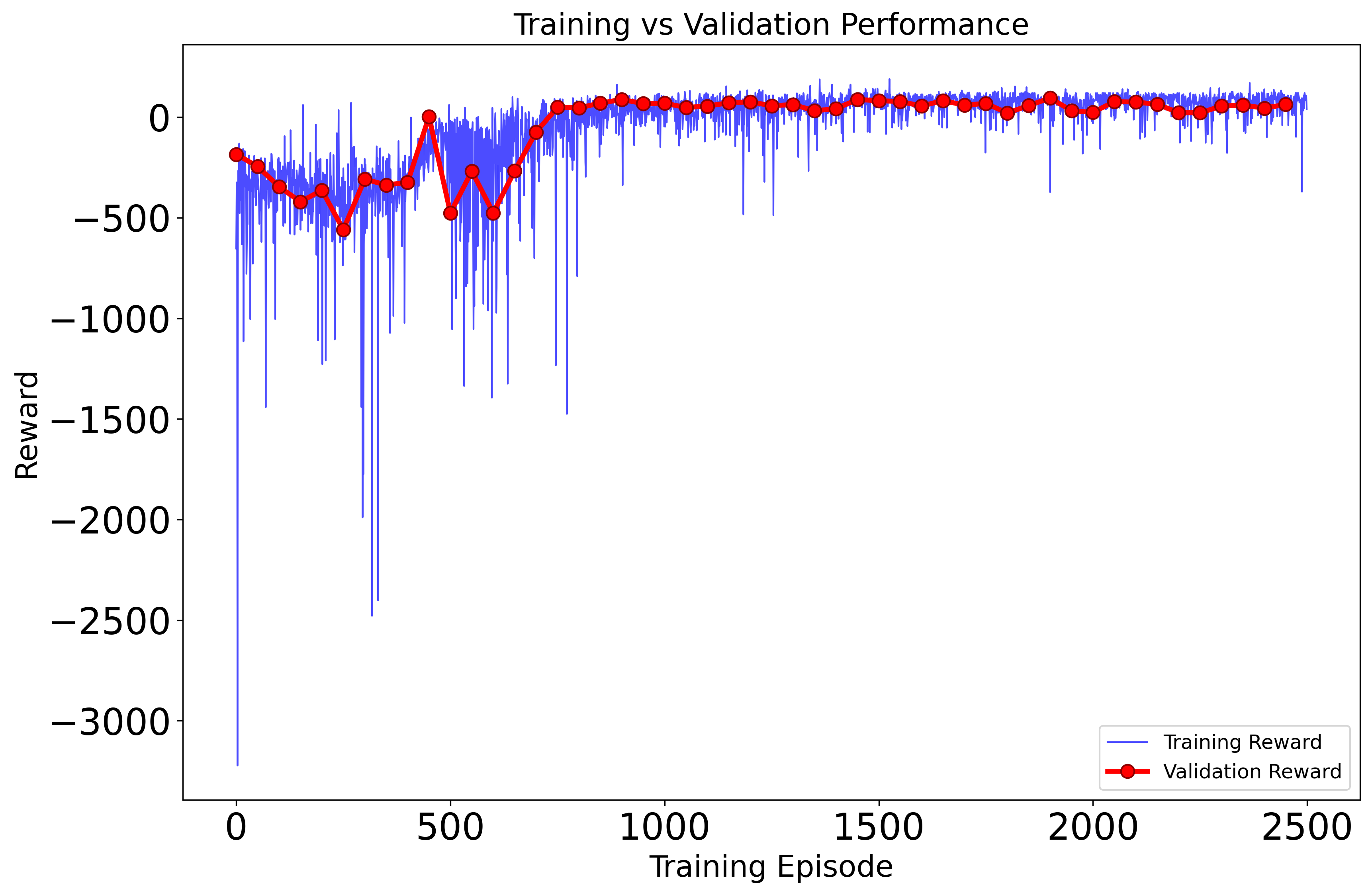}
    \caption{Training and validation episode rewards of the DDQN agent over $N_{\text{ep}} = 2500$ episodes.}
    \label{fig:ddqn-train-val}
\end{figure}

\medskip

\noindent
The evolution of the DDQN agent's return during training is illustrated in Fig.~\ref{fig:ddqn-train-val}. The blue curve shows the episode reward on the training set across all $N_{\text{ep}} = 2500$ episodes, whereas the red markers show the average evaluation reward over a small number of validation episodes at regular intervals. At the beginning of learning, the training rewards are strongly negative and exhibit large oscillations, but after a few hundred episodes they gradually increase and, around episodes 800--1000, enter a relatively stable region close to or above zero. The validation rewards follow a similar trend, after an initial transient phase, they fluctuate around a positive and approximately constant level. This behaviour indicates that, under the chosen hyper parameters, the DDQN
agent converges to a reasonably stable incentive policy on both the training
and validation sets, with no clear evidence of severe overfitting. The
remaining variability is mainly due to the stochastic nature of the
environment and the use of an $\varepsilon$-greedy exploration policy during
training.

\subsubsection{Load Reduction and Optimal Incentive Rate}

\begin{figure}[!t]
    \centering
    \begin{subfigure}[b]{0.31\textwidth}
        \centering
        \includegraphics[width=\textwidth]{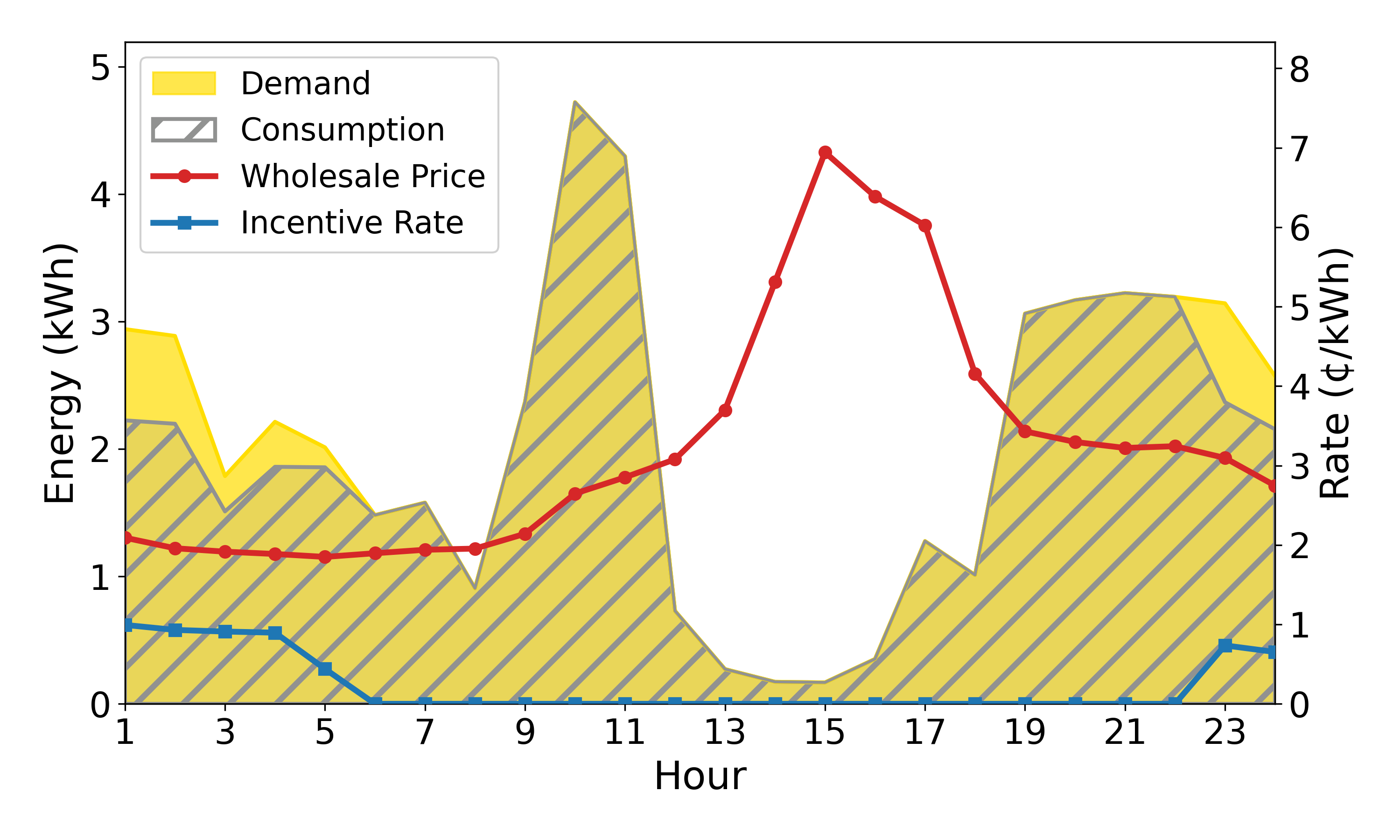}
        \caption{End User 1}
        \label{fig:load-661-b}
    \end{subfigure}
    \hspace{0.05cm}
    \begin{subfigure}[b]{0.31\textwidth}
        \centering
        \includegraphics[width=\textwidth]{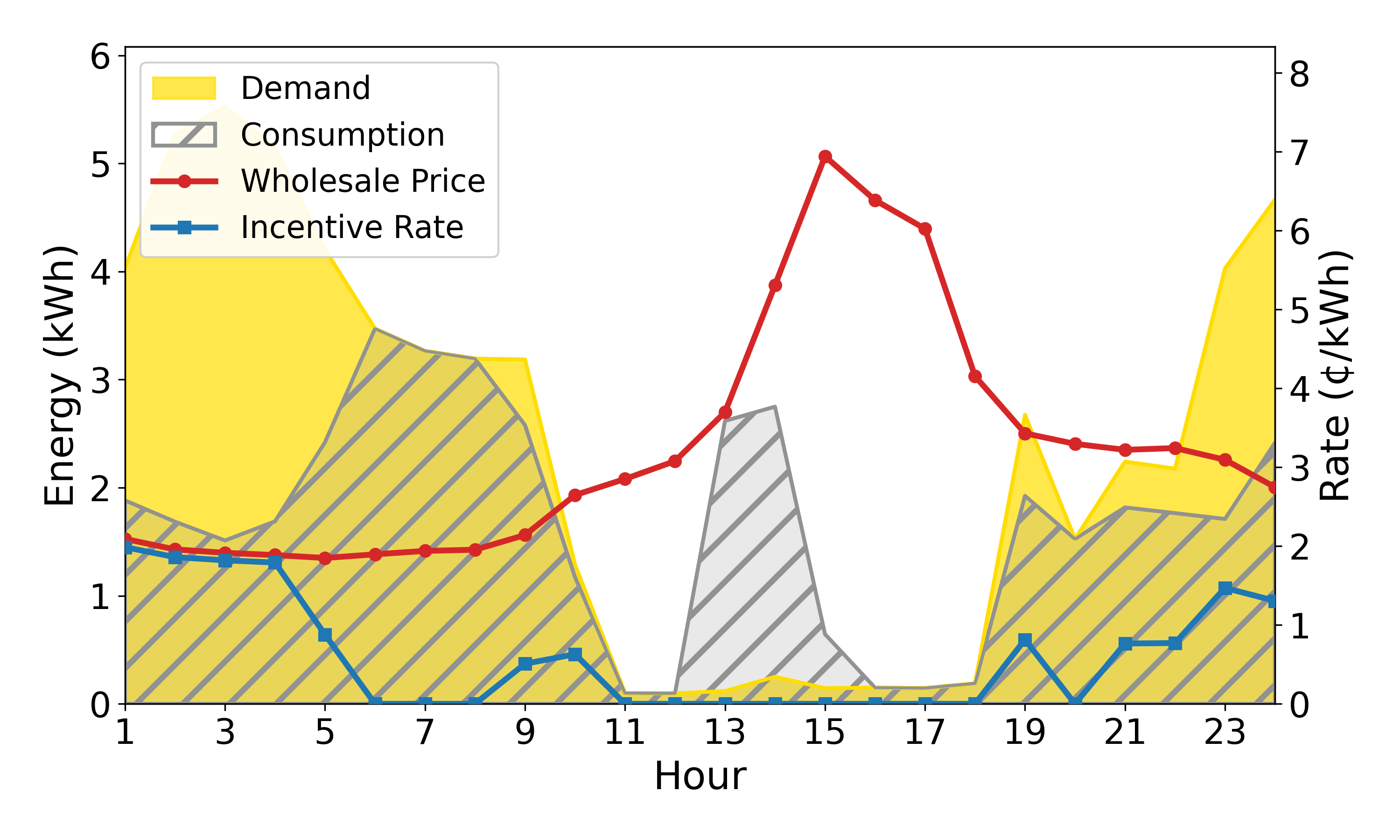}
        \caption{End User 2}
        \label{fig:load-3039-a}
    \end{subfigure}
    \hspace{0.05cm}
    \begin{subfigure}[b]{0.31\textwidth}
        \centering
        \includegraphics[width=\textwidth]{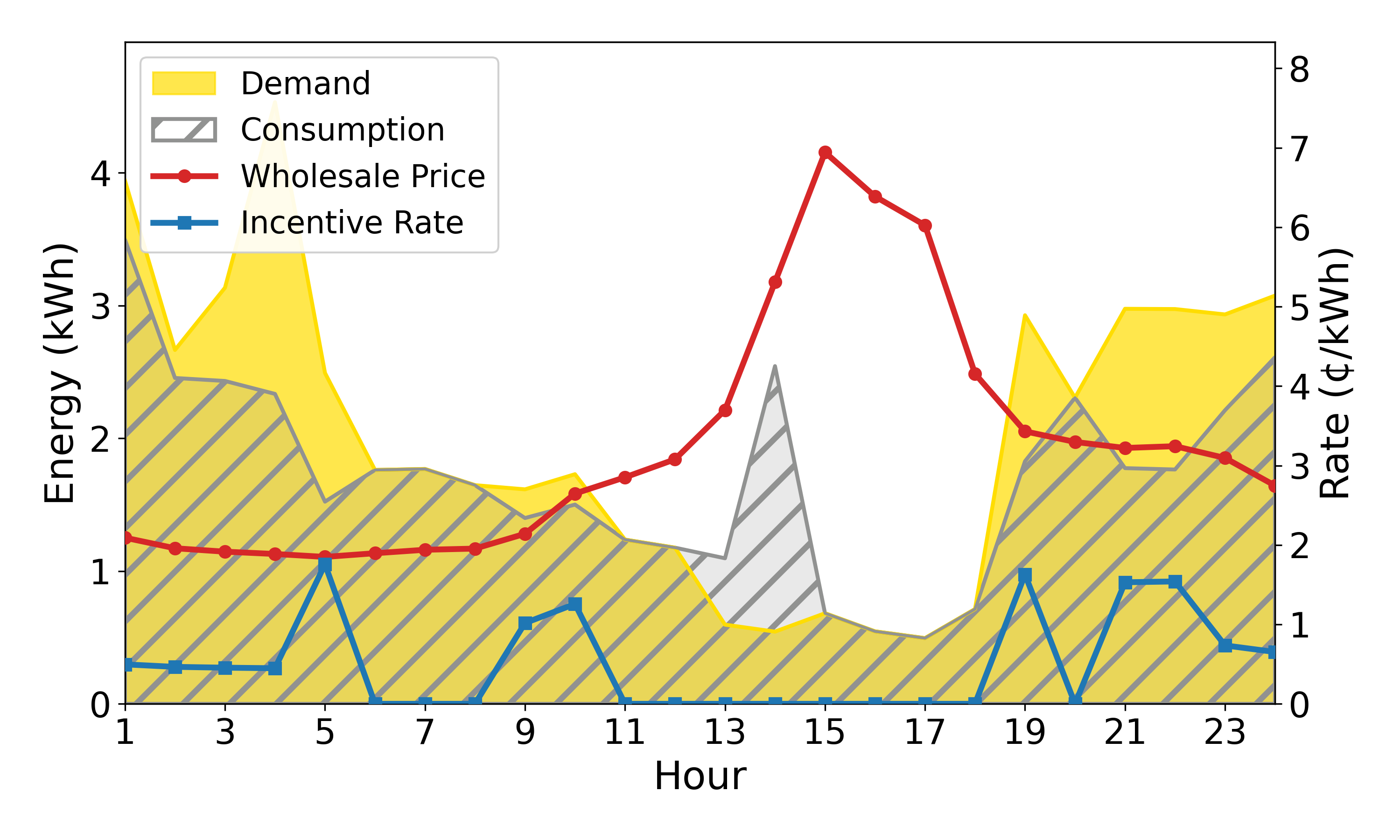}
        \caption{End User 3}
        \label{fig:load-8565-a}
    \end{subfigure}

    \caption{Results of different end users for July 27, 2018.}
    \label{fig:house-profiles}
\end{figure}

\noindent
In this part, we examine how the three residential EUs respond to the incentive signals on a representative test day. Fig.~\ref{fig:house-profiles} shows the energy profile of each EU separately. In each subfigure, the yellow area labelled ``Demand'' represents the preferred demand profile, whereas the grey hatched area labelled ``Consumption'' shows the actual consumption after applying the DR scheme. 
\medskip

\noindent
The RL agent primarily activates incentives during high-load hours when the aggregate demand approaches the capacity threshold. For some households, a portion of the demand is shifted from congested periods
to other hours of the same day with lower load. This smooths the individual
consumption profile and increases the margin to the aggregate capacity
limit. For EUs with larger dissatisfaction coefficients or a higher share of
inflexible appliances, the incentive levels remain lower and the achievable
reduction is more limited. As a result, their post-DR consumption remains
closer to the original pattern. This heterogeneous behaviour shows that the DDQN agent can tailor the
incentive rates to the characteristics of each household, including
dissatisfaction parameters and the appliance set. The required demand
reduction is therefore achieved without imposing excessive discomfort on the
participants.

\begin{figure}[!t]
    \centering
    \begin{minipage}{0.48\textwidth}
        \centering
        \includegraphics[width=\textwidth]{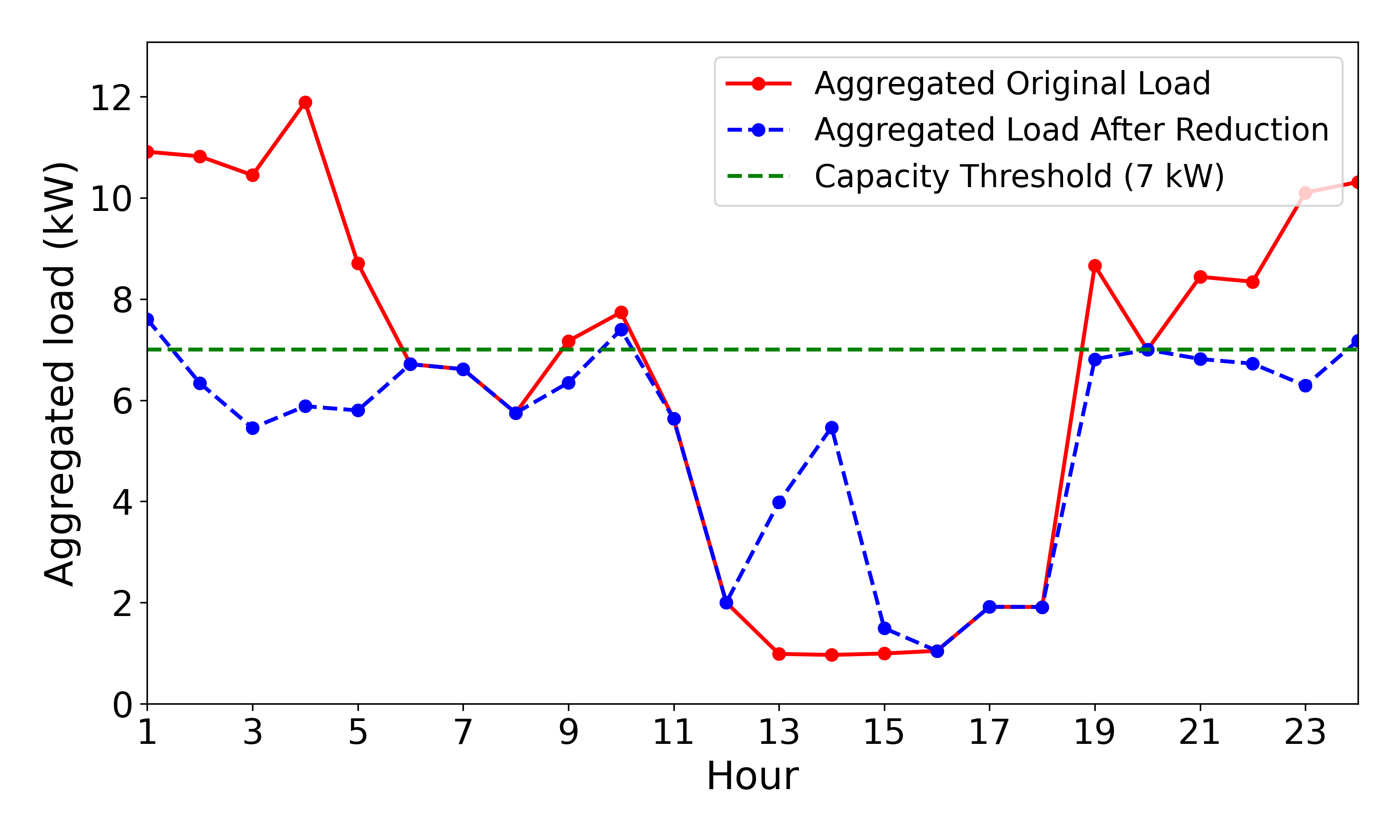}
        \caption{Aggregated load curve with capacity threshold for July 27, 2018.}
        \label{fig:agg-load-day208}
    \end{minipage}
    \hfill
    \begin{minipage}{0.48\textwidth}
        \centering
        \includegraphics[width=\textwidth]{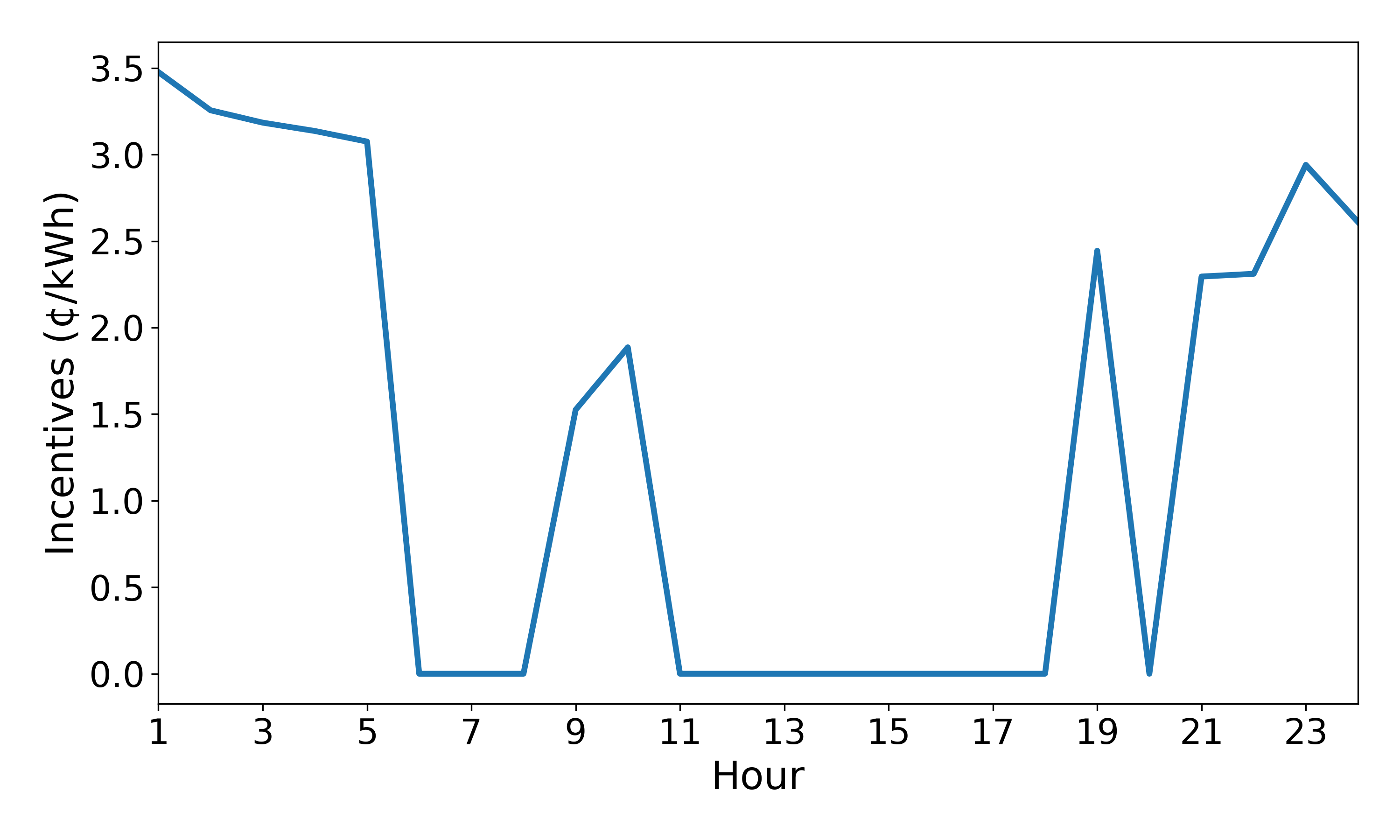}
        \caption{Total hourly incentive rate sent to the three end users on July 27, 2018.}
        \label{fig:agg-incentives-day208}
    \end{minipage}
\end{figure}

\medskip
\noindent
Next, we analyse the aggregate effect of the proposed DR scheme on the aggregated load of the three households and the way the RL agent sets the incentives for the same representative test day. Fig.~\ref{fig:agg-load-day208} shows the aggregate load profile of the three households before and after DR together with the capacity threshold. The green dashed line indicates the fixed capacity threshold of \(7~\text{kW}\), which is treated as the maximum allowable aggregate load for these three households.

\medskip

\noindent
As can be seen, at the beginning of the day and especially in the evening and
night hours, the aggregate load without DR exceeds the 7~kW capacity limit
several times. After applying the DR scheme, the main peaks are significantly
reduced, and the load profile remains close to, and mostly below the
threshold. Part of this difference is due to direct consumption curtailment in congested hours, and part is the result of shifting the demand of flexible appliances from these congested periods to other hours of the same day where the underlying load is lower. Unlike the MARL-iDR profile reported in~\cite{van2023marl}, the CCRL-DR trajectory in Fig.~\ref{fig:agg-load-day208} does not exhibit a significant rebound peak after the main peak period. In this way, both the height of the peaks is reduced and the aggregate load profile becomes smoother, without unnecessarily decreasing the total daily energy consumption across all hours.

\medskip
\noindent Fig.~\ref{fig:agg-incentives-day208} shows the total hourly incentives sent to the three households on the same day. The horizontal axis represents the hour of day and the vertical axis the aggregate incentive rate (¢/kWh). Comparing the two figures suggests that the DDQN agent has implicitly learned to account for the capacity limit on the aggregated load and to offer incentives only when they help relieve congestion through load reduction or shifting. Taken together, these two plots summarise the overall behaviour of the proposed DR system.

\medskip
\noindent
Table~\ref{tab:july-results} summarises the impact of the proposed CCRL-DR scheme when the results are averaged over all days in July. The peak of the aggregated load is reduced from $11.89~\text{kW}$ under the No-DR to $7.60~\text{kW}$ with CCRL-DR, corresponding to a peak reduction of approximately $36\%$. The mean load decreases from $6.46~\text{kW}$ to $5.34~\text{kW}$ by about $17\%$. This suggests that the DR strategy mainly shifts flexible demand to other hours, with only limited direct curtailment, rather than reducing demand uniformly across the day. As a result, the PAR improves from $1.84$ to $1.42$, corresponding to a reduction of roughly $23\%$, indicating a flatter daily load profile under the CCRL-DR scheme.

\begin{table}[H]
\centering
\caption{Results averaged per day in July}
\label{tab:july-results}
\begin{tabular}{lcc}
\hline
\textbf{Metric} & \textbf{No DR} & \textbf{CCRL-DR} \\
\hline
Peak load (kW) & 11.89 & 7.60 \\
Mean load (kW) & 6.46 & 5.34 \\
PAR            & 1.84 & 1.42 \\
\hline
\end{tabular}
\end{table}

\noindent 
\subsubsection{\texorpdfstring{Impact of the weighting factor $\rho$ on the financial analysis of the SP and EUs}{Impact of the weighting factor rho on the financial analysis of the SP and EUs}}

\medskip
\noindent
Table~\ref{tab:financial-analysis} provides a compact summary of the overall financial behaviour of the system when $\rho$ in the EU objective function is varied. For each value of $\rho$, the table reports, for each EU, the total energy reduction, incentive income, dissatisfaction cost, and net profit. For small values of $\rho$ (e.g., $\rho = 0.1$ and $\rho = 0.3$), the table shows that, for some EUs, the dissatisfaction cost is much larger than the incentive income. As a result, their net profits become negative and the sum of the three EUs' profits is also negative. Under this operating condition, the programme is not economically attractive for the EUs. Table~\ref{tab:financial-analysis} shows that higher values of $\rho$ lead to greater incentive income and lower dissatisfaction costs, and beyond a certain range all three EUs achieve a positive net profit. Under these conditions, the DR programme becomes both technically feasible and financially beneficial for the users.

\medskip
\noindent The lower part of Table~\ref{tab:financial-analysis} characterises the financial performance of the SP. For small values of $\rho$, the SP profit is high
because the revenue from load reduction is large while incentive payments
remain relatively small. Under this operating condition, the EUs experience
negative net profit.
 As $\rho$ increases, the DR cost for the SP grows and its profit decreases to some extent, but, in return, the EUs' net profits become positive and their aggregate profit increases. Overall, Table~\ref{tab:financial-analysis} shows that $\rho$ acts as a policy tuning parameter that controls the balance between the SP's benefit and the aggregate benefit of the EUs. For small values of $\rho$, the system prioritises maximising SP profit at the
expense of EU welfare. For larger values of $\rho$, the SP remains profitable,
while a larger share of the DR surplus is transferred to the EUs through
incentives, resulting in positive net profit for all parties. This result shows that the choice of $\rho$ should reflect the design objective, whether prioritising SP profit or achieving a more balanced outcome between the SP and the EUs.

\begin{table}[H]
\centering
\caption{Financial analyses of the service provider and three end users}
\label{tab:financial-analysis}
\resizebox{\textwidth}{!}{%
\begin{tabular}{l
c c c @{\hskip 12pt}
c c c @{\hskip 12pt}
c c c @{\hskip 12pt}
c c c @{\hskip 12pt}
c c c}
\hline
\textbf{Metric} & \multicolumn{3}{c}{$\rho=0.1$} & \multicolumn{3}{c}{$\rho=0.3$} & \multicolumn{3}{c}{$\rho=0.5$} & \multicolumn{3}{c}{$\rho=0.7$} & \multicolumn{3}{c}{$\rho=0.9$} \\
\hline
& EU1 & EU2 & EU3 & EU1 & EU2 & EU3 & EU1 & EU2 & EU3 & EU1 & EU2 & EU3 & EU1 & EU2 & EU3 \\
\hline
Total energy reduction (kWh) & 5.81 & 14.21 & 8.7 & 11.59 & 17.29 & 4.99 & 11.21 & 17.71 & 7.26 & 4.40 & 6.90 & 11.08 & 3.38 & 21.86 & 9.66\\
EU incentive income (¢) & 0.47 & 0.77 & 0.72 & 5.64 & 4.3 & 1.37 & 8.59 & 7.77 & 2.15 & 5.64 & 4.82 & 9.49 & 2.54 & 30.58 & 9.14 \\
EU discomfort cost (¢) & 2.14 & 91.81 & 22.37 & 6.63 & 72.90 & 16.35 & 4.39 & 52.83 & 11.92 & 1.19 & 0.87 & 8.31 & 0.14 & 10.83 & 2.65 \\
EU profit (¢) & -1.67 & -91.04 & -21.65 & -0.99 & -68.8 & -14.97 & 4.20 & -45.05 & -9.76 & 4.45 & 3.95 & 1.18 & 2.40 & 19.74 & 6.48 \\
{Sum of EU profit (¢)} & \multicolumn{3}{c}{-114.36} & \multicolumn{3}{c}{-84.56} & \multicolumn{3}{c}{-50.61} & \multicolumn{3}{c}{9.58} & \multicolumn{3}{c}{28.62} \\
SP gross income of DR (¢) & \multicolumn{3}{c}{69.34} & \multicolumn{3}{c}{79.50} & \multicolumn{3}{c}{92.14} & \multicolumn{3}{c}{55.57} & \multicolumn{3}{c}{81.90} \\
SP Cost of DR (¢) & \multicolumn{3}{c}{1.97} & \multicolumn{3}{c}{11.32} & \multicolumn{3}{c}{18.53} & \multicolumn{3}{c}{19.96} & \multicolumn{3}{c}{42.27} \\
SP profit (¢) & \multicolumn{3}{c}{67.37} & \multicolumn{3}{c}{68.17} & \multicolumn{3}{c}{73.61} & \multicolumn{3}{c}{35.61} & \multicolumn{3}{c}{39.62} \\
\hline
\end{tabular}
}
\end{table}

\begin{figure}[H]
    \centering
    \includegraphics[width=0.7\textwidth]{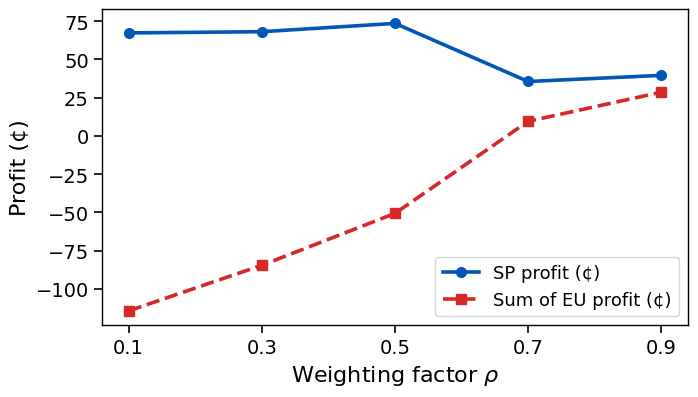}
    \caption{Sensitivity of the service provider profit and the aggregated end-users profit with respect to the weighting factor $\rho$.}
    \label{fig:sensivity_rho_profit}
\end{figure}

\medskip
\noindent To better illustrate the sensitivity of financial outcomes to the weighting factor $\rho$, Fig.~\ref{fig:sensivity_rho_profit} visualizes the variation of the SP profit and the aggregated EU profit reported in Table~\ref{tab:financial-analysis}.

\subsubsection{Comparative evaluation with an EBLR approach}
\noindent In this subsection, we compare the results of the proposed CCRL-DR method with an EBLR approach commonly used in the literature~\cite{lu2019incentive,wen2020modified, xu2020modified, lu2018dynamic}.
In this EBLR formulation, the hourly load reduction $\Delta E_{n,h}$ is
expressed in Eq.~\eqref{eq:elasticity_model}. This equation models the
reduction as a function of the demand elasticity $\xi_h$ at hour $h$ and the
position of the applied incentive rate within its allowable range. Overall,
when the incentive is closer to its maximum bound, the expected reduction
increases proportionally, while the elasticity coefficient determines the
sensitivity of the demand to those incentive changes. The representative values
of $\xi_h$ used for different periods of the day are summarised in
Table~\ref{tab:elasticity}.

\medskip
\noindent
Two key differences distinguish our CCRL-DR implementation from the EBLR
approach. First, in CCRL-DR the load reductions $\Delta E_{n,h}$ are obtained
through a device-level HEMS optimisation, whereas in EBLR,  $\Delta E_{n,h}$ is
computed directly from an aggregate elasticity model in
Eq.~\eqref{eq:elasticity_model}. Consequently, CCRL-DR captures appliance-level
flexibility, while EBLR represents EU response only at an aggregate level. Second, CCRL-DR explicitly accounts for the grid capacity limit in both the state
and reward formulation, which guides the DDQN agent to concentrate incentives
during hours when the aggregated load approaches the capacity threshold. In contrast, the EBLR approach does not incorporate an explicit capacity constraint,
and the resulting incentive–response mechanism does not explicitly adapt to actual
network congestion conditions.

\medskip
\noindent
To enable a meaningful comparison, the EBLR approach is applied to the same
three households and uses the same forecasted price and load trajectories, and time
horizon as in the CCRL-DR experiments. The elasticity
coefficients $\xi_h$ (Table~\ref{tab:elasticity}) and the EU-related parameters
$\mu_n$, $\omega_n$, $K_{\min}$, $K_{\max}$, ${\lambda_{\min}}$ and ${\lambda_{\min}}$ (Table~\ref{tab:EU-params}) are
loosely adapted from the illustrative settings in~\cite{lu2019incentive} and
have been adjusted to match the characteristics of our three-household case
study. These parameters are used only within the EBLR formulation.

\begin{equation}
\Delta E_{n,h} = E_{n,h} \cdot \xi_{h} \cdot 
\frac{\lambda_{n,h}-\lambda_{\min}}{\lambda_{\min}}
\label{eq:elasticity_model}
\end{equation}

\begin{table}[t!]
\centering
\caption{Elasticity in different hours}
\label{tab:elasticity}
\begin{tabular}{lccc}
\hline
 & Off-peak & Mid-peak & On-peak \\
 & (1--6 am, 22--24 pm) & (7--16 pm) & (17--21 pm) \\
\hline
Elasticity $\xi_h$ & 0.5 & 0.3 & 0.1 \\
\hline
\end{tabular}
\end{table}

\medskip

\begin{table}[t!]
\centering
\caption{Three end users' related parameters}
\label{tab:EU-params}
\begin{tabular}{lcccccc}
\hline
\textbf{EU} & $\mu_n$ & $\omega_n$ & $K_{\text{min}}$ & $K_{\text{max}}$ & $\lambda_{\text{min}}$ & $\lambda_{\text{max}}$ \\
\hline
EU1 & 0.3 & 0.1 & 0 & $0.3 E_{n,h}$ & $0.3\,p_{\min}$ & $p_{\min}$ \\

EU2 & 0.6 &       &   &                    &           &           \\
EU3 & 0.9 &       &   &                    &           &           \\
\hline
\end{tabular}
\end{table}

\begin{figure}[t!]
    \centering
    \begin{subfigure}[b]{0.31\textwidth}
        \centering
        \includegraphics[width=\textwidth]{Figures/energy_profile_house_661_day_208_rho0.9.png}
        \caption{End User 1}
        \label{fig:load-661-a}
    \end{subfigure}
    \hspace{0.05cm}
    \begin{subfigure}[b]{0.31\textwidth}
        \centering
        \includegraphics[width=\textwidth]{Figures/energy_profile_house_3039_day_208_rho0.9.png}
        \caption{End User 2}
        \label{fig:load-3039-b}
    \end{subfigure}
    \hspace{0.05cm}
    \begin{subfigure}[b]{0.31\textwidth}
        \centering
        \includegraphics[width=\textwidth]{Figures/energy_profile_house_8565_day_208_rho0.9.png}
        \caption{End User 3}
        \label{fig:load-8565-b}
    \end{subfigure}

    \vspace{0.3cm}

    \begin{subfigure}[b]{0.31\textwidth}
        \centering
        \includegraphics[width=\textwidth]{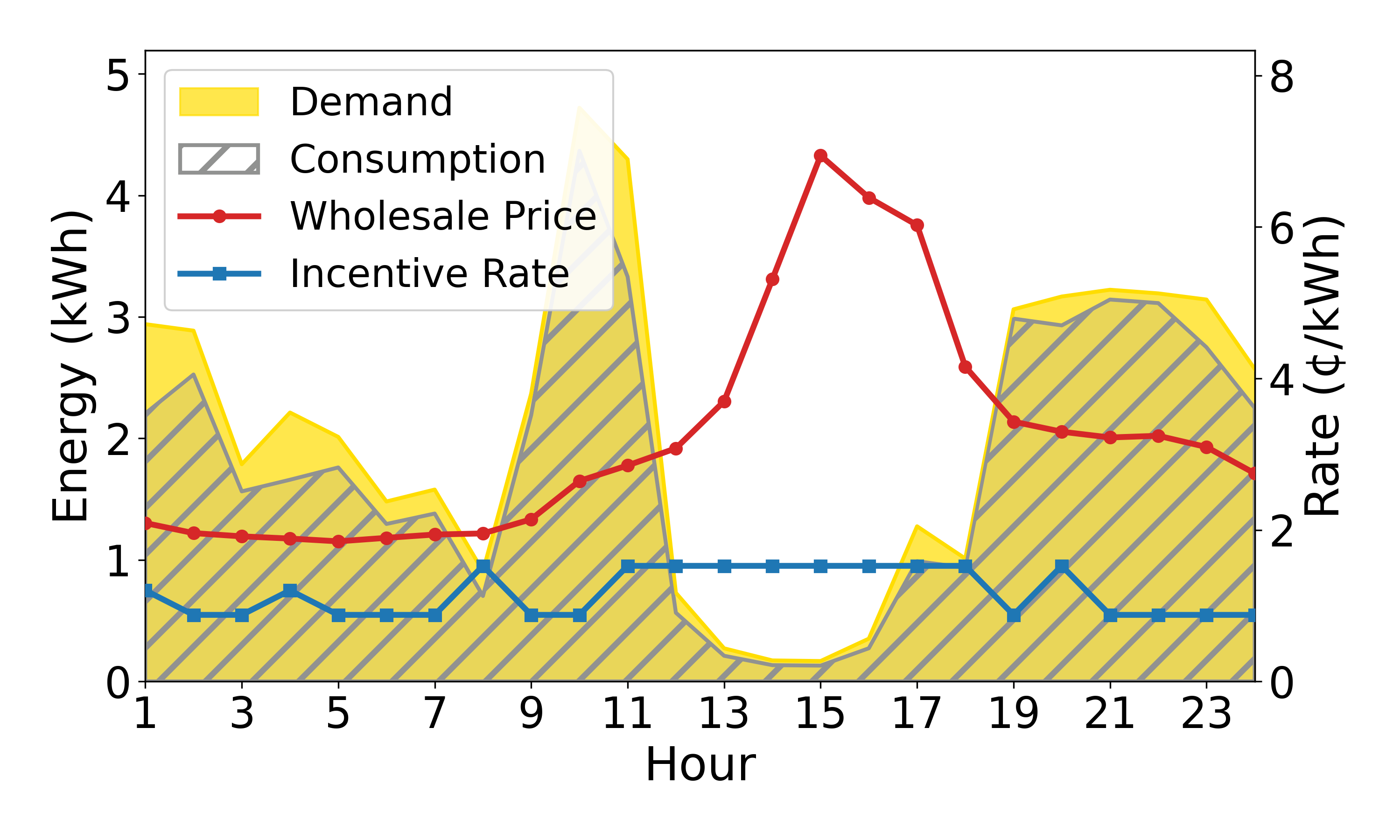}
        \caption{End User 1}
        \label{fig:alt-4031}
    \end{subfigure}
    \hspace{0.05cm}
    \begin{subfigure}[b]{0.31\textwidth}
        \centering
        \includegraphics[width=\textwidth]{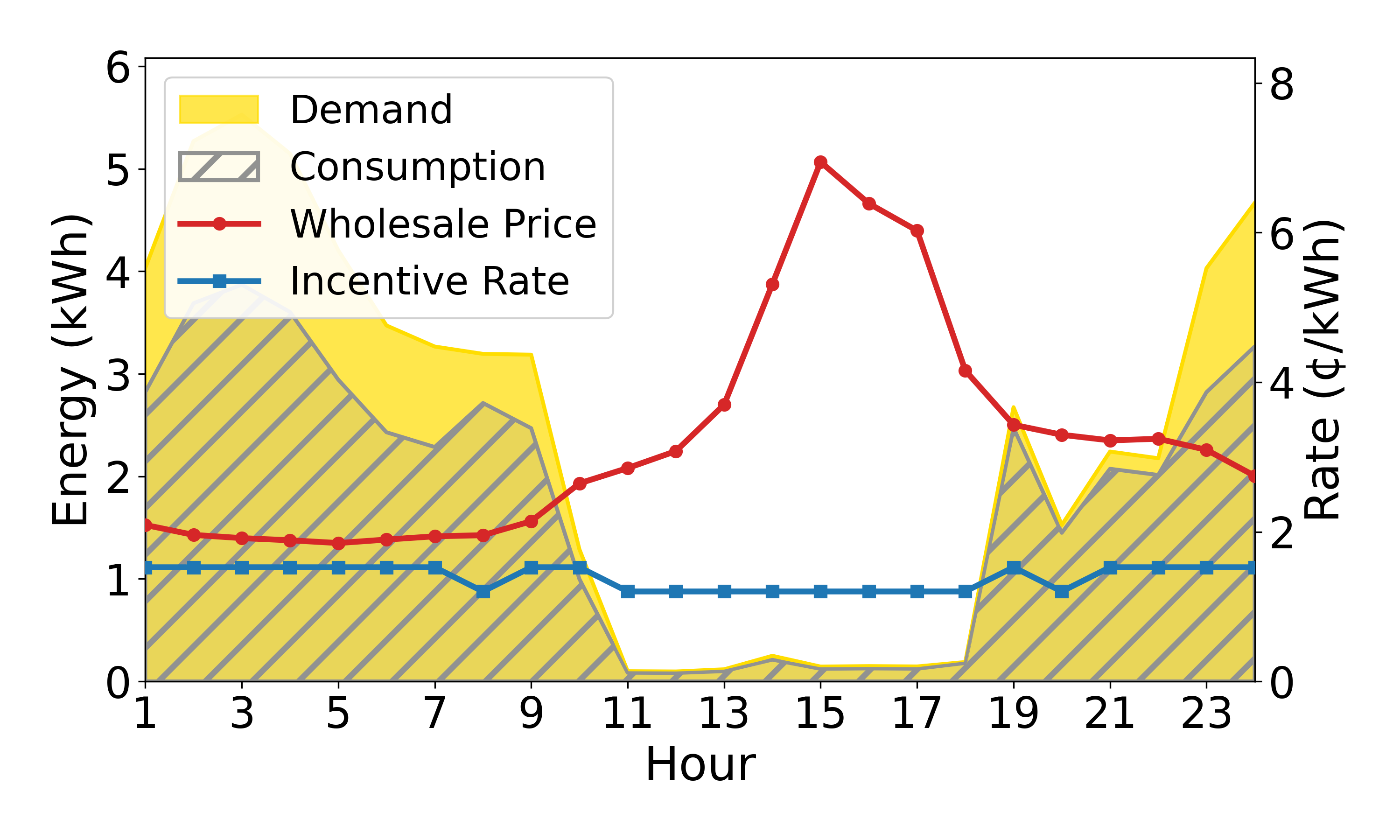}
        \caption{End User 2}
        \label{fig:alt-7800}
    \end{subfigure}
    \hspace{0.05cm}
    \begin{subfigure}[b]{0.31\textwidth}
        \centering
        \includegraphics[width=\textwidth]{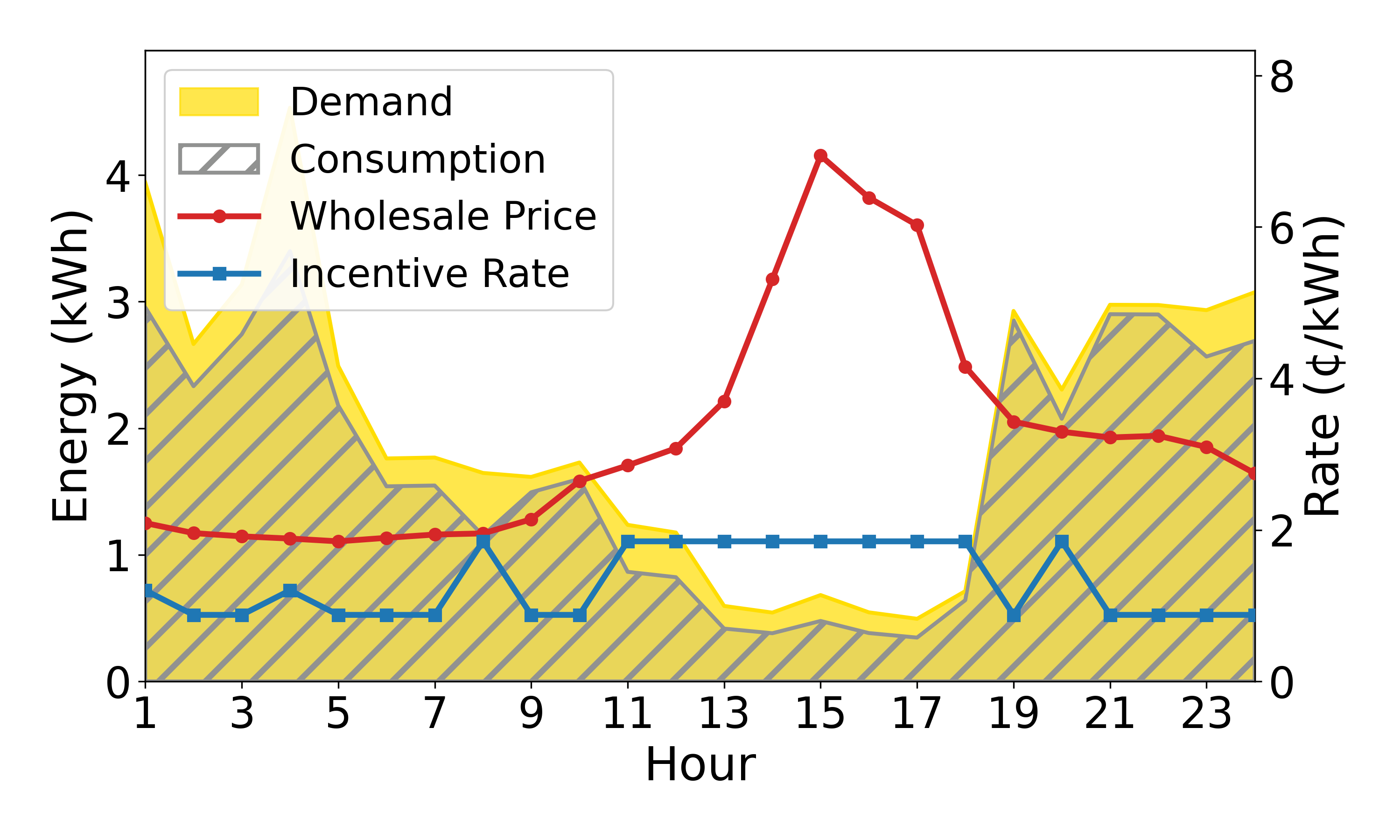}
        \caption{End User 3}
        \label{fig:alt-8156}
    \end{subfigure}

    \caption{Comparison of energy profiles for three residential end users under the CCRL-DR (top row) and EBLR  (bottom row) approaches on July 27, 2018.}
    \label{fig:comparison_proposed_alternative}
\end{figure}

\medskip
\noindent Fig.~\ref{fig:comparison_proposed_alternative} compares the hourly demand profiles of the three residential EUs on the selected summer day. In the CCRL-DR case (top row), household demand is adjusted through appliance-level responses to the incentive signals. This results in targeted peak shaving through both load curtailment and time shifting at the household level. Consequently, reductions are concentrated around local peak hours, while non-critical periods are only marginally affected. In contrast, the EBLR profiles (bottom row) are obtained from an elasticity-based EU-level response, without explicitly enforcing the network capacity constraint. As a result, the post-DR profiles exhibit more diffuse and proportional reductions, without the same degree of structure around the critical hours. Peak consumption is lowered to some extent, but the pattern of curtailment is less tightly aligned with the network congestion periods. Overall, Fig.~\ref{fig:comparison_proposed_alternative} shows that CCRL-DR exploits household heterogeneity more effectively, producing targeted demand adjustments around critical hours, whereas EBLR leads to a more uniform elastic response.

\begin{figure}[t!]
    \centering
    \begin{subfigure}[b]{0.45\textwidth}
        \centering
        \includegraphics[width=\textwidth]{Figures/aggregated_load_day_208_rho0.9.png}
        \caption{CCRL-DR approach with capacity threshold.}
        \label{fig:aggregated-proposed}
    \end{subfigure}
    \hspace{0.05cm}
    \begin{subfigure}[b]{0.45\textwidth}
        \centering
        \includegraphics[width=\textwidth]{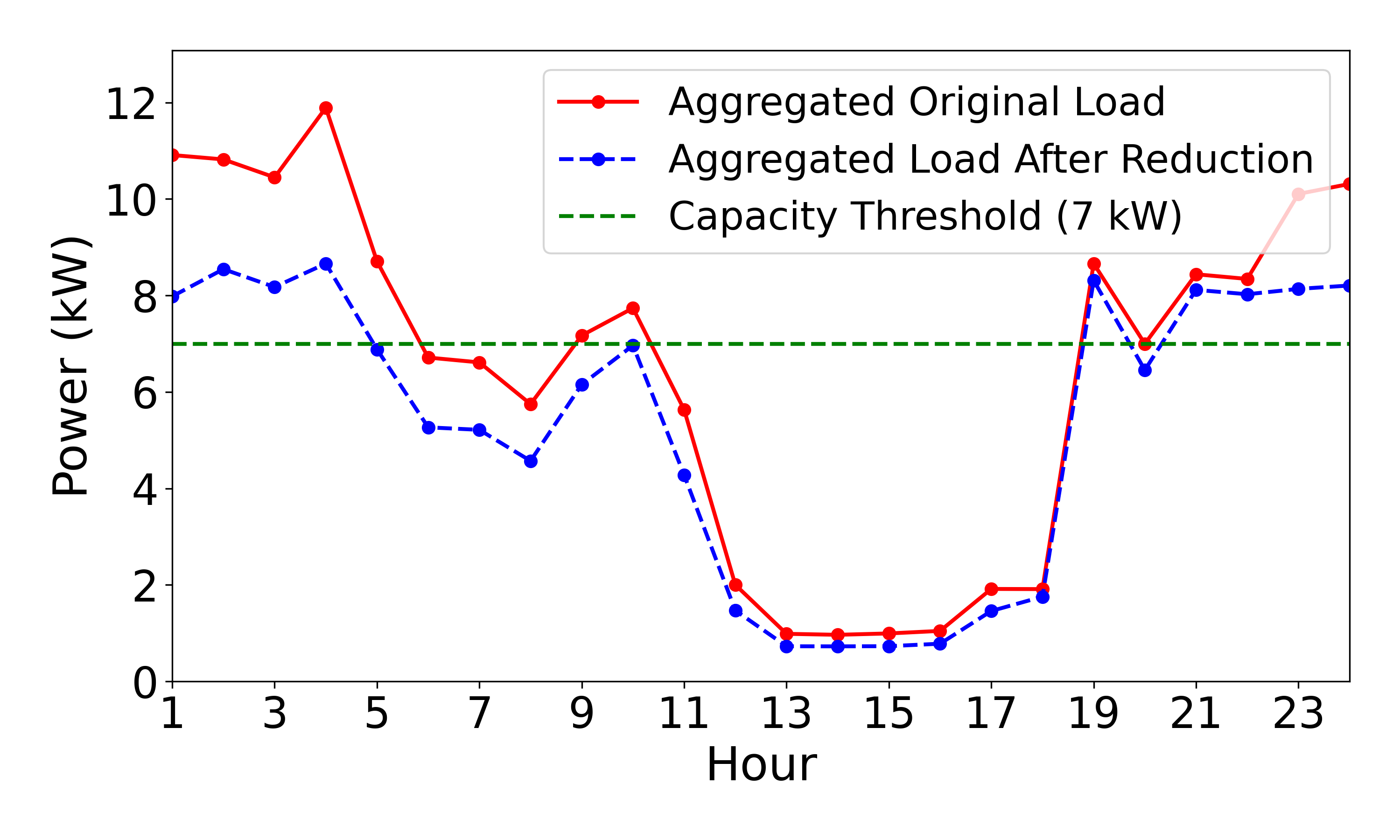}
        \caption{EBLR  approach without threshold constraint.}
        \label{fig:aggregated-alternative}
    \end{subfigure}

    \caption{Comparison of aggregated load curves for July 27, 2018 under the CCRL-DR (left) and EBLR (right) methods. The figure illustrates differences in peak load behaviour between the two approaches.}
    \label{fig:aggregated_comparison}
\end{figure}

\medskip
\noindent Fig.~\ref{fig:aggregated_comparison} moves from the household perspective to the system perspective by plotting the aggregated load of the three EUs. In the CCRL-DR case, the aggregated load is explicitly constrained by the capacity limit. This leads to a clear alignment between the incentive pattern and the periods when the aggregated load is close to the capacity limit. The aggregated post-DR curve is consistently kept at or below the capacity threshold during the critical hours, while it closely follows the original profile in non-congested periods. In other words, CCRL-DR concentrates flexibility in the most critical periods without unnecessarily reducing demand elsewhere. In the EBLR case, aggregate reductions are driven by elasticity-based responses without explicit consideration of network constraints. Although the overall load is reduced, the timing and magnitude of the reductions are driven mainly by the elasticity parameters rather than by network conditions. As a consequence, the post-DR aggregate profile under EBLR can still approach or exceed the capacity limit during peak hours, while part of the reduction occurs in less critical periods. Fig.~\ref{fig:aggregated_comparison} thus highlights the main system-level advantage of CCRL-DR; by embedding the capacity limit directly into the RL state and reward, it produces DR actions that are explicitly aware of this constraint, whereas the elasticity-based benchmark does not enforce it.

\begin{figure}[H]
    \centering
    \includegraphics[width=0.95\textwidth]{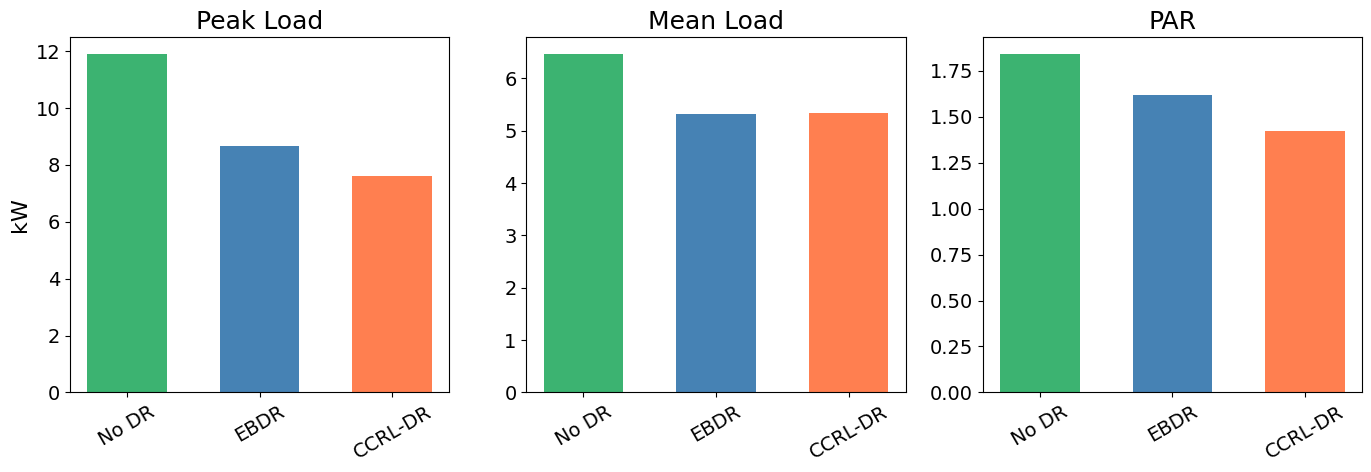}
    \caption{Comparison of average peak load, mean load, and PAR for No DR, EBLR, and CCRL-DR approaches on July 27, 2018.}
    \label{fig:avg_peak_mean_par}
\end{figure}

\medskip
\noindent Fig.~\ref{fig:avg_peak_mean_par} compares three key indicators of the aggregated load profile under three operating modes: No DR, the benchmark EBLR, and the proposed CCRL-DR. The bars show the daily peak demand, the daily mean demand, and the PAR for each case. Both DR schemes reduce the peak demand relative to the No DR, but CCRL-DR achieves the largest reduction, consistent with the behaviour observed in the aggregated load profiles. At the same time, the mean demand under CCRL-DR is only moderately lower than in the No DR, indicating that the scheme focuses on the most critical hours rather than uniformly reducing consumption across the whole day. As a result, CCRL-DR achieves the lowest PAR among the three cases, reflecting a flatter load profile that better respects the capacity limit. Overall, Fig.~\ref{fig:avg_peak_mean_par} shows that CCRL-DR achieves a more effective balance between peak reduction and energy conservation.

\noindent

\section{Conclusion and future work}

\noindent 
This paper presented a capacity-constrained incentive-based demand response
framework for residential smart grids, in which an SP employs a DRL approach
to determine real-time incentive rates under explicit grid capacity limits.
Heterogeneous EU preferences were modelled through appliance-level HEMSs
and dissatisfaction costs, enabling differentiated and adaptive incentive
design. In the considered case study, the aggregated load remains close to
the capacity threshold without inducing rebound peaks, resulting in a
$22.82\%$ reduction in the PAR compared to the no-DR case. These results
highlight the potential of combining capacity-aware incentive learning with
appliance-level flexibility to improve grid performance while maintaining
EU comfort.

\medskip
\noindent 
Future work will extend the proposed single-agent RL framework by incorporating cluster formation techniques to improve scalability and coordination across larger populations of EUs. By grouping households with similar consumption patterns or flexibility characteristics, the SP will be able to design cluster-level incentive policies while preserving appliance-level comfort modelling. This extension is expected to reduce the dimensionality of the decision space and enhance learning efficiency, while maintaining user comfort and effective capacity-aware demand response performance.

\section*{Data and Code Availability Statement}
\noindent We used publicly available datasets in this study. Residential electricity
consumption data were obtained from Pecan Street Inc., Dataport, available at:
\url{https://dataport.pecanstreet.org/}, and wholesale electricity price data were
obtained from EnergyOnline, available at:
\url{https://www.energyonline.com/}. To support reproducibility, the complete implementation and data preparation
instructions are available in our GitHub repository:
\href{https://github.com/ShafaghAPashaki/Capacity-constrained-demand-response-in-smart-grid-using-deep-reinforcement-learning}{GitHub repository}

\bibliographystyle{elsarticle-num}  
\bibliography{References}

\end{document}